\UseRawInputEncoding 
\documentclass[review,authoryear]{elsarticle}
\usepackage{lineno,hyperref}\modulolinenumbers[5]

\usepackage{appendix}
\usepackage{longtable}
\usepackage{amsmath}
\usepackage{multicol}
\usepackage{subfigure}
\usepackage{blindtext}
\usepackage{tabularx}
\usepackage{algorithm, algpseudocode}
\usepackage{enumitem}
\usepackage{rotating}
\usepackage{pdflscape}
\usepackage{adjustbox}
\usepackage{amssymb}
\usepackage{mathtools}
\usepackage{hyperref}

\newcommand{\R}{\mathbb{R}}
\DeclareMathOperator*{\argmax}{arg\,max}

\newlength{\continueindent}
\usepackage{etoolbox}
\makeatletter
\newcommand*{\ALG@customparshape}{\parshape 2 \leftmargin \linewidth \dimexpr\ALG@tlm+\continueindent\relax \dimexpr\linewidth+\leftmargin-\ALG@tlm-\continueindent\relax}
\apptocmd{\ALG@beginblock}{\ALG@customparshape}{}{\errmessage{failed to patch}}
\makeatother
\journal{Engineering Applications of Artificial Intelligence}

\biboptions{round,sort,authoryear}

\begin{document}

\algrenewcommand\algorithmicindent{1.0em}%

\begin{frontmatter}

\title{Relational Dynamic Bayesian Network Modeling for Uncertainty Quantification and Propagation in Airline Disruption Management\tnoteref{mytitlenote}}
\tnotetext[mytitlenote]{This article represents sections of a chapter from the corresponding author's completed doctoral dissertation.}

\author[1]{Kolawole Ogunsina\corref{cor1}%
\fnref{fn1}}
\ead{kolawole08@gmail.com}

\author[2]{Marios Papamichalis\fnref{fn2}}
\ead{papamixmarios@gmail.com}

\author[3]{Daniel DeLaurentis\fnref{fn3}}
\ead{ddelaure@purdue.edu}

\cortext[cor1]{Corresponding Author}
\fntext[fn1]{School of Aeronautics and Astronautics, Purdue University, United States.}
\fntext[fn2]{Department of Statistics, Purdue University, United States.}
\fntext[fn3]{School of Aeronautics and Astronautics, Purdue University, United States.}

%\newpageafter{author}

%\author[Purdue University]{Elsevier Inc}

%\author[mysecondaryaddress]{Purdue University Thesis Office\corref{mycorrespondingauthor}}
%\cortext[mycorrespondingauthor]{Kolawole Ogunsina}
%\ead{kolawole08@gmail.com}


\begin{abstract}
Disruption management during the airline scheduling process can be compartmentalized into proactive and reactive processes depending upon the time of schedule execution. The state of the art for decision-making in airline disruption management involves a heuristic human-centric approach that does not categorically study uncertainty in proactive and reactive processes for managing airline schedule disruptions. Hence, this paper introduces an uncertainty transfer function model (UTFM) framework that characterizes uncertainty for proactive airline disruption management before schedule execution, reactive airline disruption management during schedule execution, and proactive airline disruption management after schedule execution to enable the construction of quantitative tools that can allow an intelligent agent to rationalize complex interactions and procedures for robust airline disruption management. Specifically, we use historical scheduling and operations data from a major U.S. airline to facilitate the development and assessment of the UTFM, defined by hidden Markov models (a special class of probabilistic graphical models) that can efficiently perform pattern learning and inference on portions of large data sets. 

We employ the UTFM to assess two independent and separately disrupted flight legs from the airline route network. Assessment of a flight leg from Dallas to Houston, disrupted by air traffic control hold for bad weather at Dallas, revealed that proactive disruption management for turnaround in Dallas before schedule execution is impractical because of zero transition probability between turnaround and taxi-out. Assessment of another flight leg from Chicago to Boston, disrupted by air traffic control hold for bad weather at Boston, showed that proactive disruption management before schedule execution is possible because of non-zero state transition probabilities at all phases of flight operation. 
% Enter your abstract
\end{abstract}


\end{frontmatter}

%\linenumbers

\section{Introduction}
\begin{figure}[ht!]
	\centering
	\includegraphics[scale=0.25]{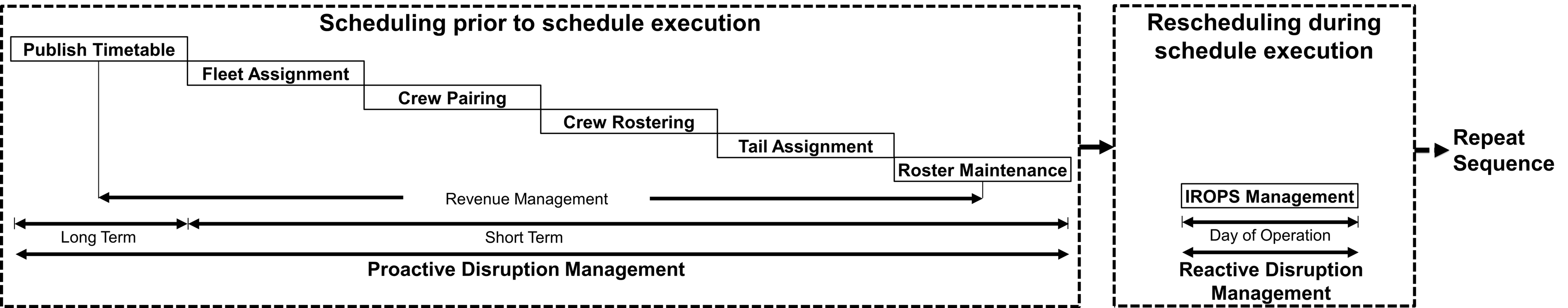}
	\caption{ The airline scheduling process}
	\label{fig:AirlineScheduling}
\end{figure}

Airlines try to maximize profit (or minimize loss) by solving problems that arise during the scheduling process shown in Fig.~\ref{fig:AirlineScheduling}. The scheduling process represents a paramount long-term and short-term planning mechanism of every airline, wherein resources (i.e. aircraft and crew) available to an airline are paired with a certain amount of passenger demand for air travel \citep{Grosche2009} that effectively define three interdependent problem dimensions: aircraft, crew, and passenger \citep{Kohl2007}. A typical airline schedule is the principal outcome of the airline scheduling process that reveals the flights offered to customers on a particular day of operation. This schedule includes assigned aircraft types, departure and arrival airports, and time of day details on how the operation of each flight unfolds from turnaround at the departure airport to aircraft gate-parking at the destination airport. 

\subsection{Irregular Operations}
Irregular operations (IROPS) are prompted by disruptive events that are likely to abound during the execution phase of the airline scheduling process depicted in Fig.~\ref{fig:AirlineScheduling}. These events which include inclement weather, equipment malfunction, and crew unavailability are most detrimental to efficiently completing airline schedule on the day of operation, because most airlines are often forced to delay (or cancel) flights in order to preserve the optimal schedule obtained prior to disruption \citep{Ball2006}. A disruption is defined as a state during the execution of an otherwise optimal schedule, where the deviation from the schedule is sufficiently large that it has to be substantially changed \citep{Galaske2016}. Airlines try to minimize unexpected costs due to disruptions (IROPS) on the day of operation by solving problems that arise during disruption management, through a few widely-accepted \textit{rules-of-thumb} implemented by human specialists in the Airline Operations Control Center (AOCC). Recent studies have revealed that disruptions yield an increased total annual operating cost of about three to five percent of the airline revenue, and airline profits would more than double if these disruptions disappeared \citep{SabreAirlineSolutions2013, Sousa2015, Amadeus2016, Gershkoff2016}. Hence, airline disruption management is the process of solving problems related to aircraft, crew and passengers when a significant deviation from the optimal schedule obtained prior to execution occurs during schedule execution on the day of operation. In that regard, reactive disruption management during schedule execution typically begins when airline scheduling for proactive disruption management prior to schedule execution ends. 
\subsection{The Problem}
From a statistical perspective, the main objective of disruption management is to eradicate the functional impact of aleatoric uncertainty \citep{Fox2011} that stems from random occurrence of disruptive events like inclement weather on optimal schedule execution on the day of operation. However, the state of the art for attaining the primary objective of airline disruption management introduces epistemic uncertainty in resolving disruptions at each phase of flight when human specialists, with different experience levels and perspectives, are required to make decisions that will affect the disruption resolution implemented in a subsequent flight phase. Although existing approaches for airline disruption management are capable of mitigating the effect of aleatoric uncertainty on scheduled flight operations, they are limited by the incapacity to explicitly address epistemic uncertainty and its impact on the quality of resolutions applied for schedule recovery and disruption management. Advancements in machine learning techniques and big data analysis \citep{C.E.Rasmussen2006,Bishop2006,Koller2009}, coupled with cost-efficient computational data storage platforms \citep{Pomerol1997}, have presented an avenue for the development and assessment of predictive and prescriptive models to facilitate the exploration of new approaches for airline disruption management that addresses the drawback of the status quo. 

Hence, we offer a robust approach that utilizes historical airline data on different \textit{rules-of-thumb} employed by human specialists in the AOCC together with current practices in airline schedule operations and recovery, to effectively quantify and minimize the propagation of epistemic uncertainty in decision-making for disruption management. 

\subsection{Contributions}
The contributions of our research are as follows:  
\begin{enumerate}
    \item We introduce an innovative uncertainty transfer function model (UTFM) architecture for concurrently tracking and assessing schedule recovery progress and decision-making during airline disruption management. The UTFM architecture relies on a relational dynamic Bayesian network (RDBN) for defining the interaction between the decision-making behavior of a characteristic human specialist (intelligent agent) in the AOCC and schedule evolution during disruption management. Thus, we integrate principles from literature on predictive analytics into literature and practices from airline schedule recovery, to enable uncertainty quantification for autonomous decision-making during disruption management.
    \item We implement a data-driven approach for executing the UTFM architecture for decision-making in airline disruption management as probabilistic graphical models. The approach utilizes historical data on schedule and operations recovery from a major U.S. airline to develop probabilistic graphical models for concurrently predicting the most likely sequence of actions for deviation from original schedule (i.e. scheduling activities) and corresponding decisions (i.e. corrective actions) enacted for disruption management during irregular airline operations. 
    \item We apply the UTFM architecture to provide an assessment of uncertainty propagation from schedule execution to schedule completion for real-world irregular schedule operations induced by weather-related disruptions. The assessment of specific real-world irregular operations on two busy routes from a major U.S. airline network revealed that decisions that resulted in significant deviations (due to bad weather) from the original schedule are most likely to occur during phases of flight operation where the aircraft is on the ground. 
\end{enumerate}
\subsection{Organization of the Paper}
We review the literature on airline schedule recovery and predictive analytics in Section \ref{literature}. Next in Section \ref{utfm_method}, we describe our UTFM architecture and discuss the data-driven and unsupervised learning approach for assembling the relational architecture by way of probabilistic graphical models. Section \ref{setup} describes our computational setup for achieving high fidelity probabilistic graphical models, while Section \ref{utfm_results} reports our results from the evaluation of actual weather-disrupted schedules from a U.S. airline by applying the UTFM architecture. We conclude with our findings and areas for further research in Section \ref{conclusion} and Section \ref{future_work} respectively.

\section{Current Practice and Literature} \label{literature}
This section provides a background of literature on airline schedule recovery during disruption management and literature on principles of predictive analytics, and how they provide suitable mechanisms to tackle existing problems in airline disruption management.

\begin{figure}[h!]
	\centering
	\includegraphics[width=0.99\textwidth]{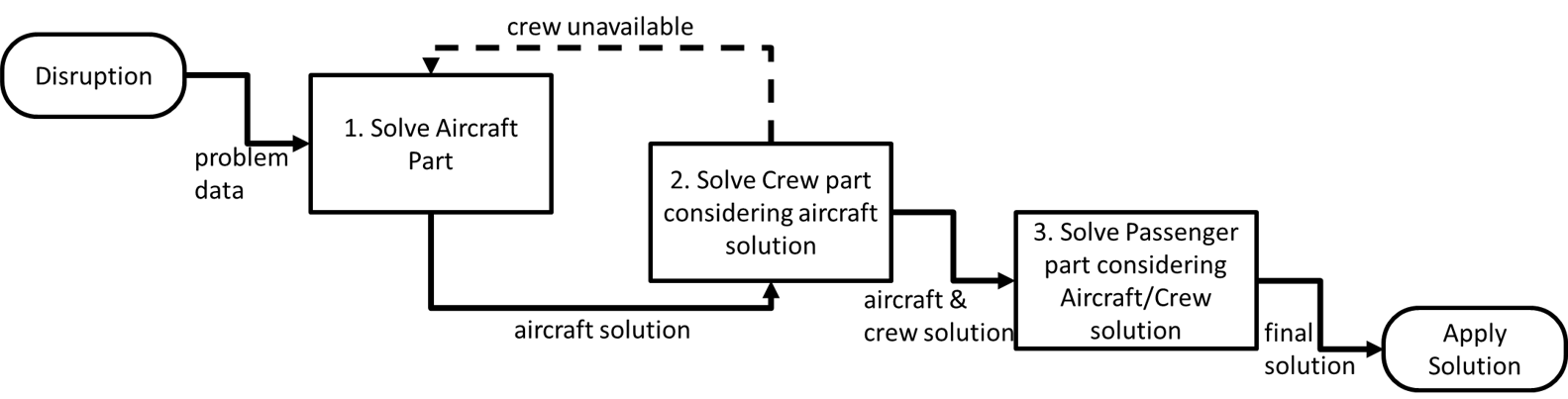}
	\caption{Current practice in airline disruption management \citep{Castro2014}}
	\label{fig:CurrentFramework}
\end{figure}

\subsection{Airline Disruption Management}
The Airline Operations Control Center (AOCC) typically addresses irregular operations in a sequential manner, such that issues related to the aircraft fleet, crew members, and passengers are resolved respectively, in that order, by their corresponding human specialists \citep{Barnhart2009}. This chronological resolution process, depicted in Fig.~\ref{fig:CurrentFramework}, is characterized by phases of flight operation \citep{Midkiff2004} where human specialists stationed in the AOCC proactively monitor and mitigate problems and disruptions related to aircraft, crew members, and passengers in the airline network during schedule execution on day of operation.

\cite{Castro2014} expressed in their work that different resolution paradigms used for airline disruption management can be categorized based upon the problem dimensions that can be readily recoverable during irregular operations. They analyzed sixty compelling research works in airline schedule recovery published between 1984 and 2014, and their findings reveal a predominance of classes of literature on solution paradigms for primarily resolving aircraft and crew dimensions (i.e. aircraft recovery and crew recovery). Compared to aircraft recovery and crew recovery, there has been relatively few published research on the integrated and simultaneous recovery of aircraft, crew, and passenger dimensions. 

While only a few decision support systems exist that jointly address two or more problem dimensions without the need for additional assessments by human specialists at each solution phase of the current recovery practice (i.e. integrated recovery), the underlying framework for developing these decision support systems and other support systems used for the airline scheduling process as a whole are monolithic, primarily based upon operations research (OR) methods (i.e. explicit optimization of goal functions in time-space networks), and often deterministic \citep{Clarke1998,Rosenberger2000,Barbati2012,Marla2017}. As such, adding supplemental features to the existing airline disruption management framework significantly increases model complexity and the computational time needed to effectively recover a disrupted airline schedule. In addition, many existing decision support systems for airline disruption management are unable to simultaneously address all the problem dimensions in airline operations, while recovering the original airline schedule, partly due to the propagation and evolution of disruptions during the operations recovery process \citep{Lee2018}.

A collaboration between the Amadeus IT group and Travel Technology Research Limited (two major IT corporations in the global travel industry) recently established that limited bandwidth of human specialists has significantly contributed to the lack of progress in developing complete and effective solutions to airline disruption management \citep{Amadeus2016}. Several key decisions at each phase of the recovery practice shown in Fig.~\ref{fig:CurrentFramework}, such as corrective actions implemented for a certain disruption type, are made by human specialists. Human specialists are flexible in decision-making, but they are not capable of accurately analyzing copious amounts of data necessary for concurrent real-time decision-making for all problem dimensions during schedule and operations recovery. Adding more personnel to the AOCC does not effectively increase human bandwidth, especially for major airlines, as network size increases \citep{Amadeus2016}. To this end, our work focuses on adopting principles from machine learning and data-driven predictive techniques for developing an architecture for robust airline disruption management. The proposed architecture enables an efficient utilization of available historical airline data for simultaneous schedule and operations recovery of all problem dimensions (i.e. simultaneously-integrated recovery).

\subsection{Predictive Analytics}
\cite{Castro2014} introduced and demonstrated the first and only published application of principles from predictive analytics in airline disruption management that enables simultaneously-integrated recovery of all problem dimensions. For an optimal schedule recovery plan, the authors use a multi-agent system design paradigm to define a model-free interaction among functional roles in the AOCC, which enabled intelligent agents to negotiate the best utility for their respective problem dimensions through the Generic Q-Negotiation (GQN) reinforcement learning algorithm \citep{Watkins1992}. Although \cite{Castro2014} provide a qualitative and
quantitative framework for discerning and modeling
adaptive decision-making for airline disruption management, their approach is statistically inefficient because the model-free environment, wherein intelligent agents interact through reinforcement learning \citep{Dayan2008}, does not employ (or estimate) a predefined flight schedule and operations model consistent with airline scheduling practices to obtain optimal disruption resolutions during airline schedule recovery. As a result, their approach requires considerable trial-and-error experience to obtain acceptable estimates of future consequences from adopting specific resolutions during airline disruption management. In contrast with the work by \cite{Castro2014}, our proposed framework leverages real-world historical data to eliminate the necessity of trial-and-error experience for facilitating simultaneously-integrated recovery during airline disruption management. 

Thus, to summarize, this paper enhances prior literature on simultaneously-integrated recovery in two major ways:
\begin{enumerate}
    \item We adeptly use experience (i.e. historical data on airline schedule and operations recovery) to construct an internal model of the transitions and immediate outcomes of scheduling activities and decisions for different phases of flight operations, by effectively describing the model environment as a relational dynamic Bayesian network architecture. The architecture defines the interaction between schedule changes and decision-making during airline disruption management, for a unique intelligent agent in a multi-agent system.
    \item We provide a modular approach for implementing an uncertainty transfer function model for disruption management. The approach inculcates feature engineering and probabilistic graphical modeling methods that enable the use of appropriate machine learning algorithms to effectively calibrate parameters for a relational dynamic Bayesian network architecture.
\end{enumerate}

\section{The Uncertainty Transfer Function Model} \label{utfm_method}
The debilitating effect of disruptions on the optimal execution of a scheduled revenue flight becomes more pronounced with increasing number of flight legs \citep{Gershkoff2016}. According to the International Air Transport Association (IATA), a scheduled revenue flight is any flight schedule executed by an airline for commercial remuneration according to a published timetable, and each flight leg in a scheduled revenue flight represents an aircraft's trip from one airport to another airport without any intermediate stops. 

\begin{figure}[h!]
	\centering
	\includegraphics[width=0.99\textwidth]{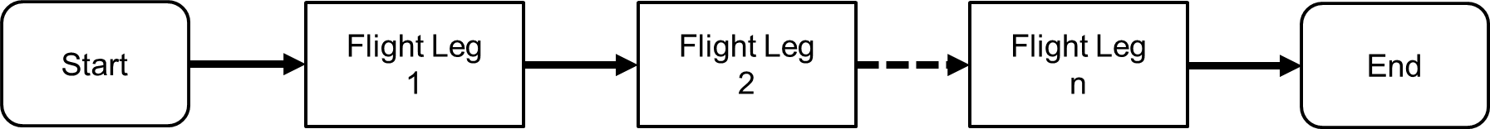}
	\caption{Disruption management for a scheduled flight defined by a Markov decision process}
	\label{fig:ADM_MDP}
\end{figure}

Every flight leg in a scheduled flight is defined by phases of aircraft activity (or flight phases) that are influenced by the decision-making activities of multiple air transportation stakeholders as the aircraft journeys between airports. For an airline, human specialists located in the AOCC perform important decision-making activities at respective flight phases during each flight leg in a scheduled flight, where actions implemented during the most precedent flight leg influence the changes in schedule and decisions made in subsequent flight legs.
Thus, fundamentally, the decision-making process for managing disruptions in a scheduled flight adheres to the Markov property \cite{Frydenberg1990}, as illustrated in Fig.~\ref{fig:ADM_MDP}. Congruently, schedule changes and decisions at a future flight phase (conditional on both past and present flight phases) during a flight leg are strictly dependent on the schedule changes and decisions made for mitigating irregular operations in the present flight phase, and not on the sequence of schedule changes and decisions made during the flight phases that preceded it. 

\subsection{Problem Formulation as a Relational Dynamic Bayesian Network} \label{formulation}
We formulate our UTFM framework \citep{Ogunsina2020} for airline disruption management as a relational dynamic Bayesian network (RDBN) \citep{Friedman1999, Sanghai2005, Getoor2007} wherein the modeling domain is defined as an airline route network containing multiple related flight schedules that are routinely executed and randomly disrupted over a certain time frame. The RDBN architecture provides a generative modeling approach that defines a probability distribution over instances of scheduled (disrupted) flights in an airline route network. By employing data features (attributes) that provide a logical description of airline activities for disruption management coupled with probabilistic graphical model templates (schema), the RDBN architecture defines the probabilistic dependencies in a domain across two time slices. Thus, for our RDBN architecture, the following general and interrelated definitions apply \citep{Sanghai2005, Neville2007, Koller2009}:

\textbf{Definition 1} (\textit{Dynamic Relational Domain})
\newline \textbf{\textit{Syntax}}: \textit{A term represents any flight phase, flight leg, or flight schedule in an airline route network domain}. 
    \textit{A predicate represents any concatenation of attributes or activities for any term in the domain.}
    \begin{itemize}
        \item \textit{The dynamic relational domain is the set of constants, variables, functions, terms, predicates and atomic formulas $Q(r_{1}, ... , r_{n}, t)$ that define an airline route network, such that each argument $r_{i}$ is a term and $t$ is the time step during disruption management.}
        \item \textit{The set of all possible ground predicates at time $t$ is determined by substituting the variables in a low-level schema of each argument with constants and substituting the functions in a high-level schema of each argument with resulting constants.}
    \end{itemize} 
    
    \textbf{\textit{Semantics}}: \textit{The state of an airline route network domain at time $t$ during disruption management is the set of ground predicates that are most likely at time $t$.}
    \newline \textbf{\textit{Assumptions}}: 
    \begin{itemize}
        \item \textit{The dependencies in an airline route network domain are first-order Markov such that ground predicates at time $t$ can only depend on the ground predicates at time $t$ or $t-1$.}
        \item \textit{A grounding (i.e. referential learning or decoding process) in an airline route network domain at time $t-1$ precedes a grounding at time $t$, such that this assumption takes priority over the ordering between predicates in the domain.
        \begin{align*}
            Q(r_{1}, ... , r_{n}, t) \prec Q(r^{\prime}_{1}, ... , r^{\prime}_{m}, t^{\prime}) \quad \text{if} \quad t < t^{\prime} 
        \end{align*}}
    \end{itemize}

\textbf{Definition 2} (\textit{Two-time-slice relational dynamic Bayesian network: 2-TRDBN}) 
\newline \textbf{\textit{Syntax}}: \textit{The 2-TRDBN is any graph (or schema) that provides a probability distribution on the state of an airline route network domain at time $t+1$ given the state of the domain at time $t$. 
\newline \textbf{\textit{Semantics}}:
    For any predicate $Q$ bounded by groundings at time $t$, we have}: 
    \begin{itemize}
        \item \textit{A set of parents $Pa(Q) = \{Pa_{1}, ... , Pa_{l}\}$, such that each $Pa_{i}$ is a predicate at time $t-1$ or $t$.} 
        \item \textit{A conditional probability model for $P(Q|Pa(Q))$, which is a first-order probability tree (or a trellis) on the parent predicates.} 
    \end{itemize} 
\textbf{\textit{Assumptions}}:
    \begin{itemize}
        \item \textit{If $Pa_{i}$ is at time $t$, then $Pa_{i} \prec Q$ or $Pa_{i} = Q$.}
        \item \textit{If $Pa_{i} = Q$, then its groundings are bounded to those that precede the defined grounding of $Q$.}
    \end{itemize}

\textbf{Definition 3} (\textit{Relational Dynamic Bayesian Network: RDBN}) 
\newline \textbf{\textit{Syntax}}: \textit{A RDBN for disruption management is any network pair $(\mathcal{N_{0}}, \mathcal{N_{\rightarrow}})$, such that $\mathcal{N_{0}}$ is a dynamic Bayesian network (DBN) at time $t=0$ and $\mathcal{N_{\rightarrow}}$ is a 2-TRDBN.}
\newline \textbf{\textit{Semantics}}: \textit{$\mathcal{N_{0}}$ characterizes the probability distribution over a relational (airline route network) domain prior to schedule execution (i.e. at $t=0$). Given the state of the relational domain at a time $t$ during disruption management (or schedule execution), $\mathcal{N_{\rightarrow}}$ represents the transition probability distribution on the state of the domain at time $t+1$}. 
\newline \textbf{\textit{Assumptions}}: \textit{A term (node) is created for every ground predicate and edges are added between a predicate and its parents at a time $t > 0$.}
    \begin{itemize}
        \item \textit{Parents are obtained from $\mathcal{N_{0}}$ if $t = 0$, else from $\mathcal{N_{\rightarrow}}$.} 
        \item \textit{The conditional probability distribution for each term is defined by a probabilistic graphical model bounded by a specific grounding of the predicate.}
    \end{itemize}

\begin{figure}[ht!]
	\centering
	\includegraphics[width=0.99\textwidth]{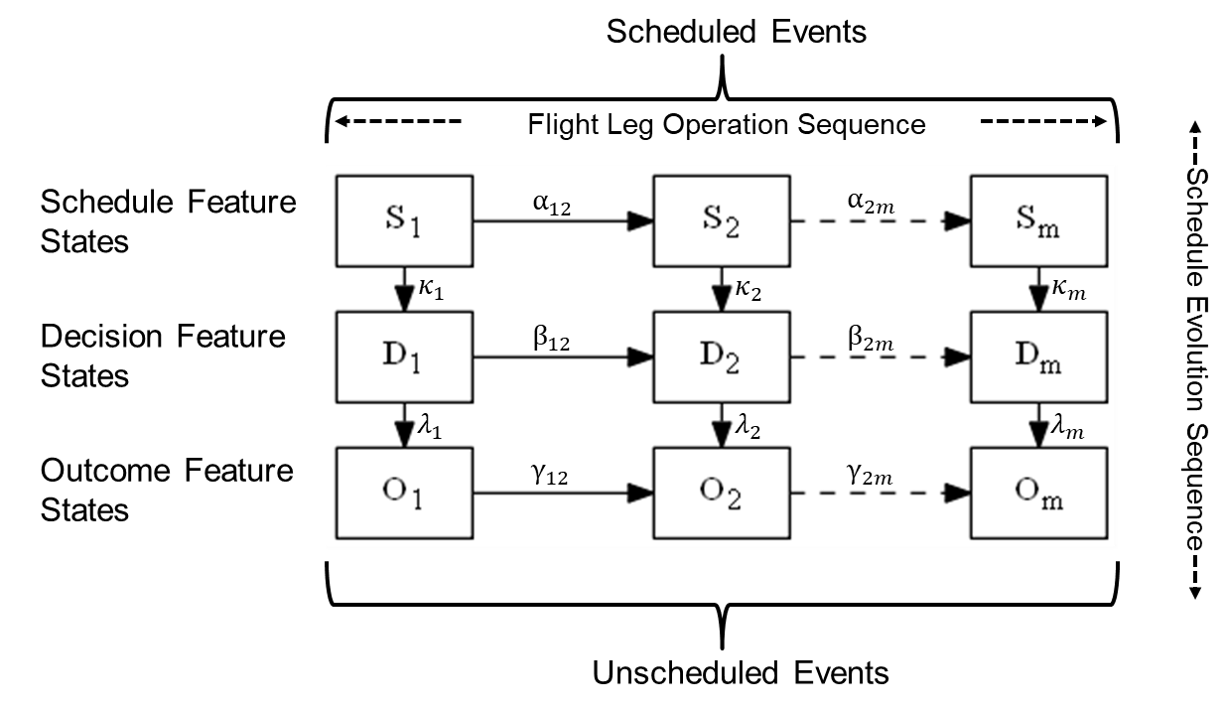}
	\caption{RDBN architecture for a representative flight leg}
	\label{fig:UncertaintyModel}
\end{figure}

For the purposes of uncertainty quantification and propagation discussed in this paper, we adapt the aforementioned definitions for a RDBN to construct a UTFM, such that the modeling domain is for a representative flight leg that is defined by the probabilistic graphical model (i.e. atomic formula) illustrated by Fig.~\ref{fig:UncertaintyModel}. The flight leg operation sequence (i.e. disruption progression along horizontal axis in Fig.~\ref{fig:UncertaintyModel}) represents the spatiotemporal axis in a multidimensional Markov chain \citep{Ching2008} that describes the order in which (or when) random disruptions (i.e. indeterminate features for bad weather events) occur during different phases of flight. As such, the flight leg operation sequence defines the propagation of aleatoric uncertainty in schedule and operations recovery. The schedule evolution sequence (i.e. schedule-planning evolution along the vertical axis in Fig.~\ref{fig:UncertaintyModel}) captures epistemic uncertainty in decision-making for operations recovery by characterizing the order in which (or how) the flight schedule changes with respect to disruption resolutions, such as \textit{rules-of-thumb} or decision features like delay period, are applied by human specialists on the day of operation. Scheduled events constitute data features (such as departure times, arrival times, aircraft type, etc.) that define the optimal airline (flight) schedule for $m$ different flight phases prior to schedule execution.

Furthermore, scheduled events serve as start points in the UTFM architecture and may also inform the decision-making of human specialists during the resolution of a specific type of disruption. Unscheduled events represent an updated set of data features that characterize the adjustment of optimal flight schedule by human specialists based upon the impact of disruption at $m$ different flight phases during schedule execution. Unscheduled events provide end points in the UTFM architecture. Schedule feature states (labeled $S$ in Fig.~\ref{fig:UncertaintyModel}) represent functions of data items that are strictly subject to uncertainty in determinate data features with respect to airline planning and scheduling prior to schedule execution. Decision feature states (labeled $D$ in Fig.~\ref{fig:UncertaintyModel}) represent functions of action items that human specialists implement during schedule execution to resolve disruptions in the optimal schedule obtained prior to schedule execution (e.g. delay time, flight swap flag, etc.), while outcome feature states (labeled $O$ in Fig.~\ref{fig:UncertaintyModel}) represent functions of data items that human specialists use to assess the impact of their decisions after resolving deviations from the optimal schedule obtained prior to schedule execution. The parameters for $S, D, O, \alpha, \beta, \gamma, \kappa, \lambda$ in Fig.~\ref{fig:UncertaintyModel} are obtained by grounding via hidden Markov models, to determine the schedule evolution and decision-making proclivities of human specialists at each flight phase during disruption management for a characteristic flight leg. Refer to algorithms in \ref{alg:Baum-Welch}, \ref{alg:Viterbi}, \ref{alg:UTFMlearning}, and \ref{alg:UTFMdecoding} for more information on UTFM grounding. 

\subsection{Data Abstraction and Feature Engineering} \label{Data_Eng}
Prior to learning and assembling the UTFM to enable the prediction of uncertainty propagation patterns during airline disruption management, it is imperative to understand the nature of the airline data set that will be used to develop constituent models. By following the atomic formula from Fig.~\ref{fig:UncertaintyModel} and appraising a raw data set defined by over 40 separate data features provided by a major U.S. airline, this section describes the methods used to abstract and encode different data features in the data set to achieve high fidelity probabilistic graphical models. 

\subsubsection{Data Abstraction}
We apply a combination of event abstraction and uncertainty abstraction principles \citep{Ogunsina2021} to establish three separate classes of data features for uncertainty quantification and propagation in airline disruption management, namely: 

\begin{enumerate}
    \item \textit{Determinate aleatoric features}: These are flight schedule features that are subject to the least possible uncertainty for the risk of alteration during irregular operations for disruption management, based upon inherent randomness of disruptive events. For instance, longitude and latitude coordinates that provide specific geographical information for origin and destination stations are always assumed to remain unchanged, by the AOCC, during the recovery of a delayed flight schedule. 
    
    \item \textit{Indeterminate aleatoric features}: These are data features that are subject to the most possible uncertainty for the risk of instantiating irregular operations during schedule execution, due to inherent randomness of disruption events. Examples include IATA delay codes that indicate inclement weather at a particular airport, which may require a human specialist in the AOCC to delay the departure of a specific flight at the (origin) airport and reassign some or all of its passengers to another flight with a later departure.     
    
    \item \textit{Epistemic features}: These are flight schedule features that are subject to the most possible uncertainty for the risk of alteration during irregular operations for disruption management, due to lack of knowledge of the exact impact of their alteration.  For instance, following a specific disruption like late arrival of flight crew for a scheduled flight, a human specialist in the AOCC may choose to delay the departure of the flight by a specific period of time after the original departure time. However, most times, the human specialist can not guarantee that the decision to apply a particular delay duration after scheduled departure will produce a specific disruption management outcome, due to the cascading effect of disruptions in large airline networks. 
\end{enumerate}

\subsubsection{Feature Engineering}

\begin{table}[ht!]
\begin{center}
\caption{Feature Engineering and Transformation for UTFM} \label{tab:data_proc}
\begin{tabularx}{\linewidth}{ 
  | >{\centering\arraybackslash}X
  | >{\centering\arraybackslash}X 
  | >{\centering\arraybackslash}X 
  | >{\centering\arraybackslash}X | }
  \hline
	\textbf{Raw Data Class} & \textbf{\textit{First-Degree} Transformation} & \textbf{\textit{Second-Degree} Transformation} & \textbf{Refined Data Type}\\ \hline
	\textit{Geographic Features} & Spherical directional vectors, geodesic distance
     & Standardization & Continuous \\ \hline
	\textit{Temporal Features} & Periodic (Sine/Cosine) vectors
     & Standardization & Continuous \\ \hline
	\textit{Categorical Features} & One-hot encoding & Standardization & Continuous\\ \hline
	\textit{Continuous Features} & N/A & Standardization & Continuous \\ \hline
\end{tabularx}
\end{center}
\end{table}
 
Since many algorithms for learning probabilistic graphical models perform best with continuous data \citep{Getoor2007, Ogunsina2019b, Ogunsina2021}, it is necessary to encode all values of features (or fields) in the data set into functional and relevant continuous data for use in appropriate algorithms \citep{Liskov1988, ReidTurner1999}. Table~\ref{tab:data_proc} reveals the feature engineering methods applied to transform the features in a raw data set for developing and assessing the UTFM. As shown in Table~\ref{tab:data_proc}, first-degree transformation represents the conversion of different attributes that define data features into appropriate mathematical functions and values, while second-degree transformation represents the conversion of data features into suitable statistical distributions based upon the number of available flight schedules (i.e. data samples). As such, raw geographical features are converted into spherical directional vectors and geodesic distance \citep{T.Vincenty1975} while raw temporal features are converted into sine or cosine vectors during first-degree transformation. Categorical data features in the raw data set are converted into sparse matrices during first-degree transformation through one-hot encoding \citep{Seger2018}. All data features (fields) are subsequently scaled to obtain a standard normal distribution during second-degree transformation to facilitate statistical interpretation of the results obtained from probabilistic graphical models. A complete definition of all the refined data features used for creating the probabilistic graphical models discussed in this paper can be found in \ref{table:detalea_table}, \ref{table:indetalea_table}, and \ref{table:epis_table}.

\subsection{Solution Approach for UTFM} \label{solution_method}

\begin{figure}[ht!]
	\centering
	\includegraphics[width=0.6\textwidth]{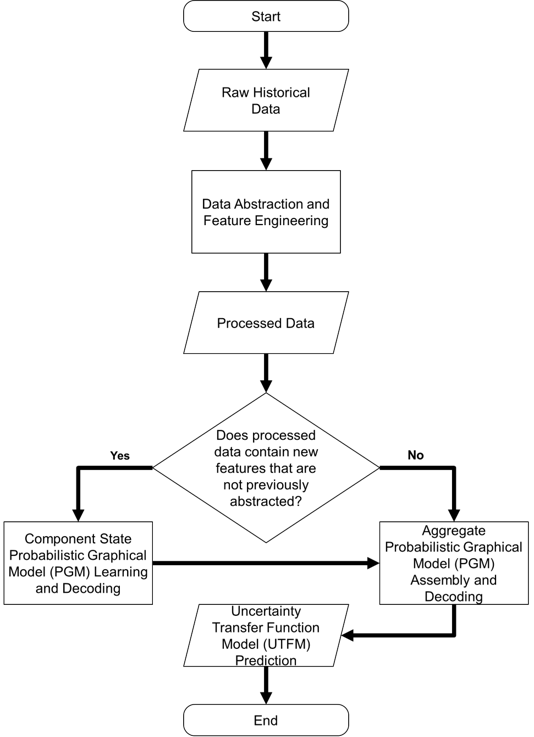}
	\caption{Component assembly approach for automatic uncertainty quantification for disruption management}
	\label{fig:UncertaintyModel_Solution}
\end{figure}

We use a solution technique based upon a component assembly process, which enables generative programming for probabilistic graphical models \citep{Koller2009}, to calibrate (ground) the parameters of the multidimensional Markov chain that define the UTFM introduced in Section \ref{utfm_method}. Component assembly is a widely espoused modeling paradigm in computer science and software engineering \citep{Cao2005}, and facilitates the integration of state components of the UTFM that define separate phases of flight operation and schedule-recovery evolution in the UTFM architecture. Through generative programming \citep{Chase2005, Czarnecki2005}, highly customized and optimized intermediate parameters defining each state component and aggregate UTFM parameters, can be created on demand from elementary and reusable parameters of state components, through a priori knowledge of the graph structure of the Markov system. 

Fig.~\ref{fig:UncertaintyModel_Solution} reveals our solution approach to automatic uncertainty quantification for airline disruption management. The approach starts by abstracting historical airline schedule and operations recovery data into a digestible data set, via the methods described in Section \ref{Data_Eng}, applicable to algorithms for predictive analytics. Next, the refined data set is used to learn optimal probabilistic graphical model parameters of each state component of the UTFM, before constructing an overarching probabilistic graphical model from the aggregation of the respective optimized probabilistic graphical models of state components.

For the remainder of this section, we introduce probabilistic graphical modeling and discuss the role of hidden Markov models for grounding (i.e. calibrating the parameters) in a probabilistic graphical model representation of the UTFM. 
\subsubsection{Probabilistic Graphical Modeling}

\begin{figure}[ht!]
	\centering
	\includegraphics[width=0.99\textwidth]{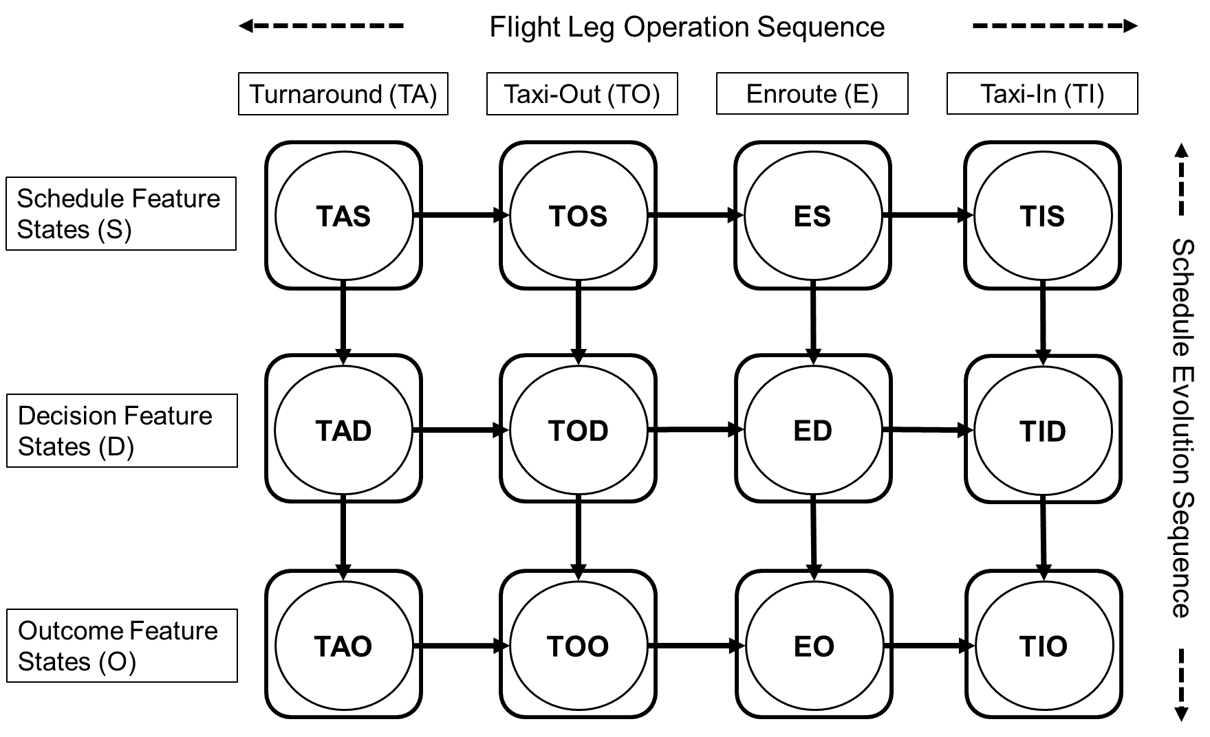}
	\caption{Probabilistic graphical model representation of UTFM}
	\label{fig:PGM_UTFM}
\end{figure}

Probabilistic graphical modeling provides an avenue for a data-driven approach to constructing the UTFM architecture, which is very effective in practice \citep{Koller2009}. By employing rudimentary activity guidelines from human specialists in the AOCC for airline disruption management, critical components for constructing an intelligent system such as representation, learning, and inference can be readily inculcated in the UTFM.

Fig.~\ref{fig:PGM_UTFM} shows the probabilistic graphical model representation of the UTFM defined by four major phases of flight along the operation sequence axis namely: Turnaround, Taxi-Out, Enroute, and Taxi-In, while the schedule evolution sequence axis is defined by three separate phases of schedule changes with respect to airline planning on day of operation namely: Schedule, Decision, and Outcome. Thus, the graph structure of the UTFM comprises of 12 distinct component states (nodes) with 12 internal state transitions and 17 external state transitions, such that each component state contains a set of combination (interaction) of data features, listed in Section \ref{setup}, that encode the behavioral proclivities of human specialists at different phases of activity during airline disruption management.

Schedule state components (i.e., TAS, TOS, ES, TIS) in Fig.~\ref{fig:PGM_UTFM} represent an interaction of data features that describe the evolution of original (optimal) flight schedule predetermined prior to schedule execution on day of operation, which would inform the decision-making of a human specialist in the AOCC during schedule execution. As such, schedule state components in the UTFM encapsulate epistemic uncertainty in proactive disruption management prior to schedule execution (i.e., uncertainty in tactical disruption management). Decision state components in the UTFM (i.e., TAD, TOD, ED, TID) define the interaction of data features that describe the action items that human specialists implement for resolving specific types of disruption that occur during schedule execution, and define epistemic uncertainty in reactive disruption management during rescheduling on day of operation (i.e., uncertainty in operational disruption management). Outcome state components in Fig.~\ref{fig:PGM_UTFM} (i.e., TAO, TOO, EO, TIO) represent the interaction of a set of data features that characterize the original schedule adjusted based upon the impact of disruption resolutions (i.e. action items) implemented by human specialists during schedule execution, and therefore define epistemic uncertainty in proactive disruption management for future airline scheduling after schedule execution (i.e., uncertainty in strategic disruption management).  

\subsubsection{Hidden Markov Models for Probabilistic Graphical Modeling of UTFM} \label{HMM_UTFM}
The hidden Markov model (HMM), also known as a transducer-style probabilistic finite state machine \citep{Vidal2005}, is the simplest class of dynamic Bayesian networks and a useful tool for representing probability distributions over a sequence of observations \citep{Ghahramani2001, Letham2012}. The hidden Markov model obtains its name from defining two separate but related characteristics. First, it assumes that the observation at a particular instance in time was generated by an arbitrary process whose state is hidden from the observer. Second, it assumes that the state of this hidden process satisfies the Markov property. To that effect, the hidden Markov model lends an appropriate grounding medium for solving the learning and inference (decoding) problems \citep{Yang1997} for the probabilistic graphical model representation and construction of the UTFM. 

Mathematically, the hidden Markov model is defined as a stochastic process $(X_{k}, Y_{k})_{k\geq0}$ on the product state space $(E\times F, \mathcal{E}\otimes \mathcal{F})$ if there exist transition kernels $P:E \times \mathcal{E} \rightarrow [0,1]$ and $\Phi:E \times \mathcal{F} \rightarrow [0,1]$ such that 

\begin{equation}\label{eqn:hmm1} 
    \begin{aligned}
        \textbf{E}(g(X_{k+1}, Y_{k+1})|X_{0},Y_{0},...,X_{k},Y_{k}) = \int g(x,y)\Phi(x,dy)P(X_{k},dx)
    \end{aligned}
\end{equation}
and a probability measure $\mu$ on $E$ wherein
\begin{equation}\label{eqn:hmm2} 
    \begin{aligned}
        \textbf{E}(g(X_{0},Y_{0})) = \int g(x,y)\Phi(x,dy)\mu(dx)
    \end{aligned}
\end{equation}
for any bounded and measurable function $g:E \times F \rightarrow \R$. As such, $\mu$ represents the initial measure, $P$ is the transition kernel, and $\Phi$ represents the observation kernel of the hidden Markov model $(X_{k}, Y_{k})_{k\geq0}$.

\subsubsection{HMM Learning} \label{HMM_learning}
The learning problem for construction of the UTFM is representative of optimizing the parameters of the pair of dynamic Bayesian networks $(\mathcal{N_{0}}, \mathcal{N_{\rightarrow}})$ defined in Section~\ref{formulation} based upon available data, and therefore presents two separate learning sub-problems: \textit{Intra-State} HMM learning and \textit{Inter-State} HMM learning. Hence, \textit{Intra-State} HMM learning and \textit{Inter-State} HMM learning characterize the grounding process for obtaining optimal parameters for $\mathcal{N_{0}}$ and $\mathcal{N_{\rightarrow}}$ respectively. Specifically, \textit{Intra-State} HMM learning represents the ability to effectively determine appropriate interaction patterns (i.e. transition likelihood) for \textit{hidden} data features (subject to epistemic uncertainty) which are embedded in each state component of the UTFM shown in Fig.~\ref{fig:PGM_UTFM}, based upon observing data features (i.e. observations) that are strictly subject to uncertainty from determinate or indeterminate aleatoric features observed at any phase of activity during airline disruption management. Some examples of data features that represent observations for \textit{Intra-State} HMM learning of state components in the UTFM include total distance between origin airport and destination airport, and total number of passengers (i.e. demand for air travel) available for flight before and after schedule execution. Thus, the primary objective of \textit{Intra-State} HMM learning is to achieve an optimal HMM (probability distribution mixture model) that is capable of efficiently predicting the likelihood of remaining at a particular phase of activity (i.e. state component) in the UTFM for airline disruption management. 

\textit{Inter-State} HMM learning, on the other hand, characterizes the ability to ascertain the interaction or transition patterns between any two neighboring state components (phases of activity) in the UTFM, wherein data features (listed in Section \ref{setup}) embedded in the state component at the future (posterior) phase of activity in the UTFM are set as observations while data features embedded in the state component at the current (prior) phase of activity are set as hidden states. As such, the primary objective of \textit{Inter-State} HMM learning is to attain an optimal HMM (probability distribution mixture model) that is capable of accurately predicting the likelihood of transitioning between present and future phases of activity (i.e. state components) in the UTFM.  

\begin{figure}[b!]
	\centering
	\includegraphics[width=0.9\textwidth]{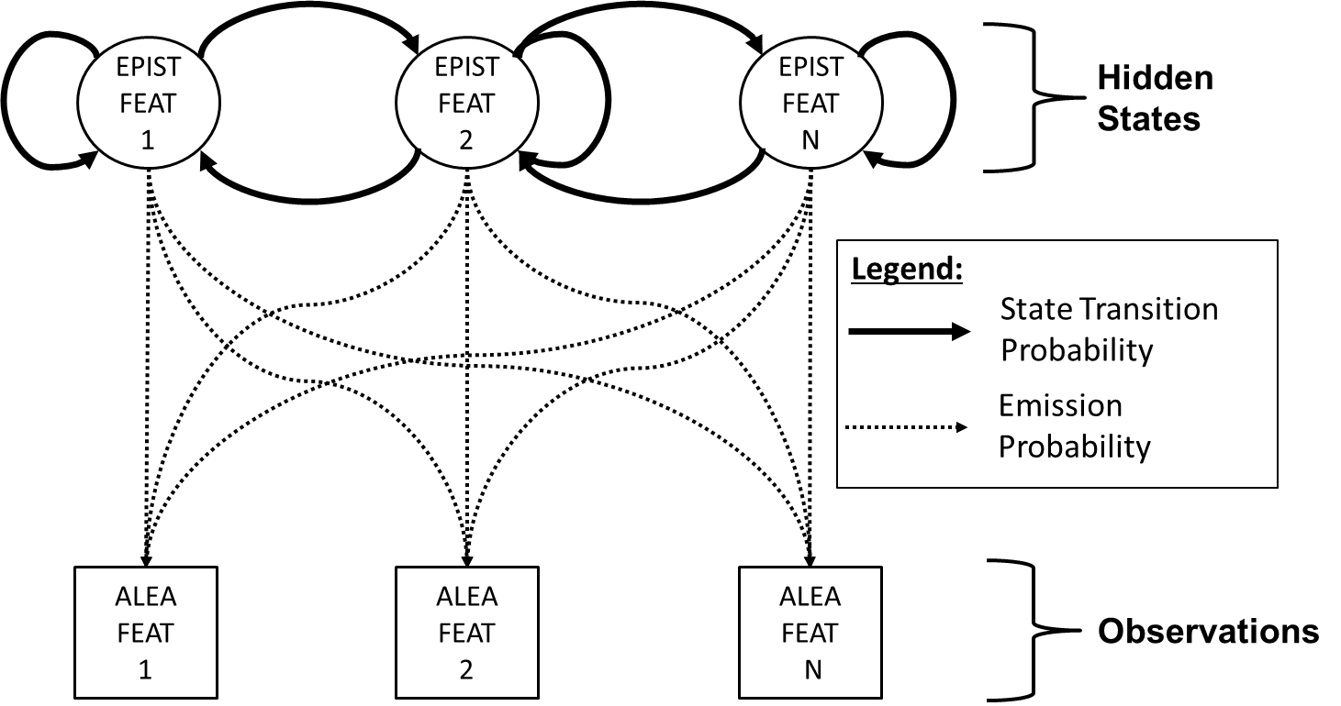}
	\caption{Intra-state HMM schema for remaining in an activity phase in UTFM}
	\label{fig:Intra_State_HMM}
\end{figure}

\begin{enumerate}
\item Compute
\begin{equation} \label{eq6}
\\ Q(\theta, \theta^\prime) = \sum_{z=\bar{Z}} log[P(X,z;\theta)]P(z|X;\theta^\prime)\\\
\end{equation}
\item Set
\begin{equation} \label{eq7}
\\\theta^{\prime+1} = \argmax_{\theta} Q(\theta, \theta^\prime)\\
\end{equation}
\end{enumerate}

The Baum-Welch algorithm \citep{Baum2007} is a dynamic programming approach that uses the expectation maximization (EM) algorithm \citep{Bilmes1998} to find the maximum likelihood estimate of the parameters of an HMM given a set of observations. The Baum-Welch algorithm presents a convenient means for learning the optimal parameters (i.e. state transition and emission probabilities) of an \textit{Intra-State} or \textit{Inter-State} HMM, because it guarantees that the optimal parameters of the HMM are easily estimated in an unsupervised manner during training by utilizing unannotated observation data \citep{Boussemart2012}.  In essence, the Baum-Welch algorithm described by steps in Equations~\ref{eq6} and ~\ref{eq7}, where $X$, $\bar{Z}$, and $\theta$ are the latent state space, observation space, and initial HMM parameters respectively, is an iterative procedure for estimating $\theta^\prime$ until convergence, such that each iteration of the algorithm is guaranteed to increase the log-likelihood of the data. However, convergence to a global optimal solution is not necessarily guaranteed \citep{Baum2007}.

\begin{figure}[b!]
	\centering
	\includegraphics[width=0.9\textwidth]{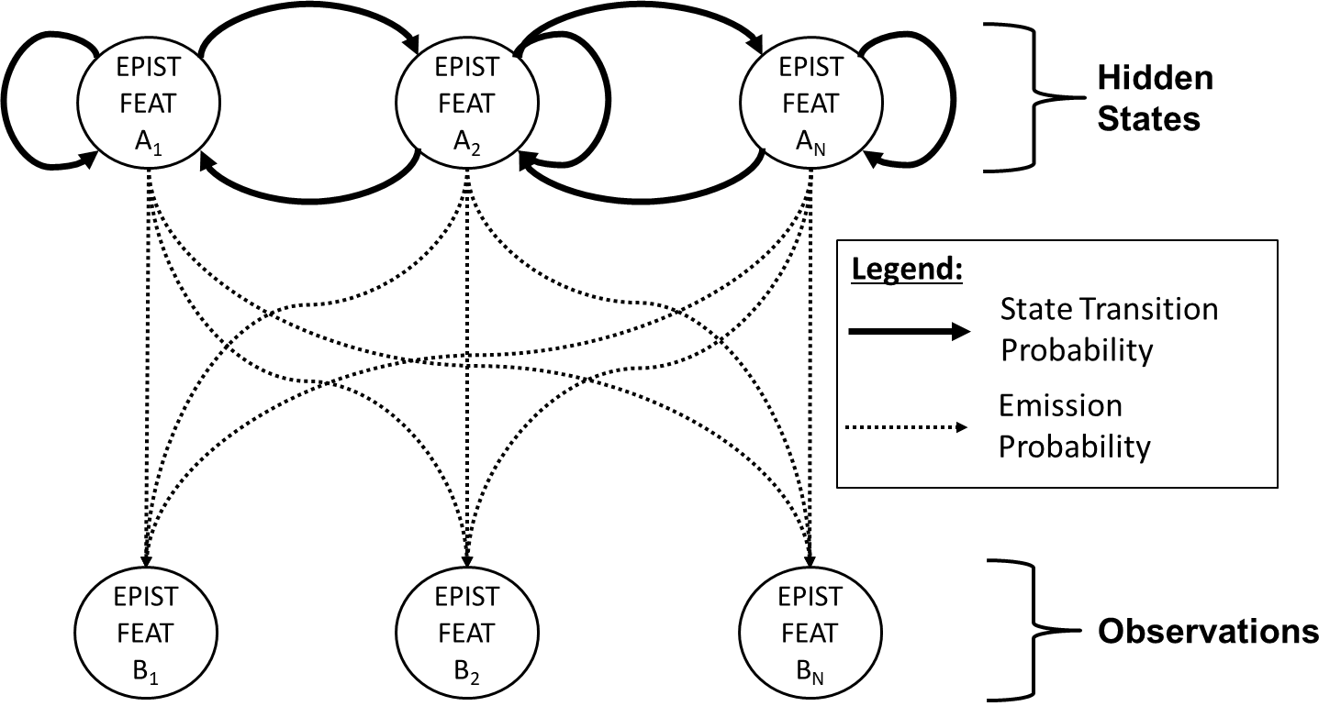}
	\caption{Inter-state HMM schema for transitioning between activity phases in UTFM}
	\label{fig:Inter_State_HMM}
\end{figure}

Fig.~\ref{fig:Intra_State_HMM} reveals the general schema for learning the optimal parameters of an \textit{Intra-State} HMM. The circles and squares in Fig.~\ref{fig:Intra_State_HMM} represent the hidden (latent) states (i.e. data features subject to epistemic uncertainty) and observations (i.e. data features which are representative of aleatoric uncertainty) respectively. The learning objective for the \textit{Intra-State} HMM schema in Fig.~\ref{fig:Intra_State_HMM} is to use the Baum-Welch algorithm to find the optimal HMM parameters, which are the solid and dashed arrows that represent state transition probabilities and emission probabilities respectively. 

Fig.~\ref{fig:Inter_State_HMM} shows a generic schema for learning the optimal parameters of a typical \textit{Inter-State} HMM, essential for predicting the likelihood of transitioning from one activity phase to another activity phase across both spatiotemporal axes in the UTFM. The circles labeled `EPIST FEAT A' and `EPIST FEAT B' in the \textit{Inter-State} HMM schema, shown in Fig.~\ref{fig:Inter_State_HMM}, represent epistemic data features embedded in current activity phase $A$ (i.e. hidden states) and future activity phase $B$ (i.e. observations), respectively, in the UTFM. Similar to the \textit{Intra-State} HMM, the learning objective for the \textit{Inter-State} HMM schema depicted by Fig.~\ref{fig:Inter_State_HMM} is to use the Baum-Welch algorithm to find the optimal HMM parameters, which are the solid and dashed arrows that represent the state transition probabilities and emission probabilities respectively.

Unlike the \textit{Intra-State} HMM schema where hidden states represent data features subject to epistemic uncertainty for disruption management and observations represent data features subject to aleatoric uncertainty, both hidden states and observations in the \textit{Inter-State} HMM schema are representative of data features subject to epistemic uncertainty for disruption management. Thus, the overarching objective of an optimal \textit{Intra-State} HMM is to accurately and expeditiously quantify the epistemic uncertainty at a specific phase of activity in the UTFM, while the overall objective of an optimal \textit{Inter-State} HMM is to precisely predict the propagation of epistemic uncertainty between different phases of activity in the UTFM, for robust airline disruption management. 

\subsubsection{HMM Inference} \label{HMM_inference}
Upon learning optimal parameters of \textit{Intra-State} and \textit{Inter-State} hidden Markov models, which define proactive and reactive behavioral patterns of human specialists at different stages of airline disruption management in the UTFM, it is imperative to conduct inference on the models to complete the assembly of the UTFM for effectively predicting uncertainty propagation patterns for airline disruption management. Similar to the learning problem, the inference problem for the assemblage of the UTFM is defined by two separate sub-problems: \textit{component} UTFM decoding and \textit{aggregate} UTFM decoding. \textit{Component} UTFM decoding, defines the capacity of both \textit{Intra-State} and \textit{Inter-State} hidden Markov models for obtaining the most probable sequence of hidden (epistemic) data features in both types of HMMs, based upon (aleatoric or epistemic) observation data features necessary for decoding in their respective schema illustrated in Figs.~\ref{fig:Intra_State_HMM} and~\ref{fig:Inter_State_HMM}. Thus, the primary objective of \textit{component} UTFM decoding problem is to provide the maximum likelihood estimates of the most probable sequence of hidden data features from optimal \textit{Intra-State} and \textit{Inter-State} HMMs upon inputting appropriate observation data features.

\textit{Aggregate} UTFM decoding, on the other hand, describes the ability of the amalgamation of all \textit{Intra-State} and \textit{Inter-State} HMMs that constitute the UTFM, to precisely estimate the quantification and propagation of epistemic uncertainty at all phases of activity in the UTFM, based upon observing the maximum likelihood estimates of the most probable sequence of hidden data features retrieved from optimal \textit{Intra-State} HMMs in the UTFM by way of \textit{component} UTFM decoding. As such, a complementary objective of \textit{aggregate} UTFM decoding problem is to obtain the parameters for $S, D, O, \alpha, \beta, \gamma, \kappa, \lambda$ as shown in Fig.~\ref{fig:UncertaintyModel}, by estimating the weighted average of the maximum likelihood estimates of the most probable sequence of hidden data features retrieved from all optimal \textit{Intra-State} and \textit{Inter-State} HMMs upon observing their respective input data features (i.e. observations).

\begin{equation} \label{eq8}
\\x^{*} = \argmax_x P(z, x | \theta^\prime)\\
\end{equation}

The Viterbi decoding algorithm \citep{Viterbi1967, Forney1973} is a proven dynamic programming algorithm that performs the HMM inference of the most probable sequence of hidden states (and its corresponding likelihood) based upon a specific sequence of observations, ultimately solving both the \textit{component} and \textit{aggregate} UTFM decoding sub-problems respectively. In principle, the Viterbi decoding algorithm defined by Equation~\ref{eq8}, where $x$, $z$, and $ \theta^\prime$ represent a sequence of hidden states, a sequence of observations, and an arbitrary HMM respectively, uses a recursive (backtracking search) procedure for obtaining the optimal sequence of hidden states from the total number of possible sequences of hidden states for a specific sequence of observations, by selecting the sequence of hidden states that has the highest probability based upon maximum likelihood estimations from the arbitrary HMM \citep{Forney1973}. As such, the Viterbi decoding algorithm provides an efficient method for avoiding the explicit enumeration of all possible combinations of sequences of hidden states (i.e. concatenations of data features) while identifying the optimal sequence (i.e. Viterbi path) of hidden states with the highest probability of occurrence or least uncertainty \citep{Omura1969}.

In summary, from a UTFM assemblage perspective, the underlying objective of \textit{component} UTFM decoding is to perform inference on all optimal \textit{Intra-State} and \textit{Inter-State} HMMs that define the UTFM, by implementing the Viterbi decoding algorithm to effectively estimate the likelihood (Viterbi probability) of the most likely sequence of hidden states (data features) based upon observing appropriate data features (observations), as shown in Figs.~\ref{fig:Intra_State_HMM} and~\ref{fig:Inter_State_HMM}. By extension, the overall objective of \textit{aggregate} UTFM decoding is to apply the Viterbi algorithm for determining the most likely sequence of state components that describes the propagation of epistemic uncertainty at different phases of activity in the UTFM shown in Fig.~\ref{fig:PGM_UTFM}. The state transition parameters of a representative probabilistic finite state machine for the UTFM are weighted averages of the Viterbi probabilities obtained via \textit{component} UTFM decoding that satisfy the properties of a stochastic matrix \citep{Haggstrom2002}.

\section{Computational Setup and Analysis} \label{setup}
We now discuss the computational framework for generating state components of the probabilistic graphical model representation of the UTFM (shown in Fig.~\ref{fig:PGM_UTFM}), which is used to predict epistemic uncertainty propagation during decision-making for airline disruption management. Prior to implementing the Baum-Welch and Viterbi algorithms to learn and decode useful HMMs for determining authentic likelihoods of internal and external transitions amongst different state components in the UTFM, raw historical airline data, necessary for enabling the application of algorithms for the development of these probabilistic graphical models, is first refined by following the data abstraction and feature engineering guidelines described in Section \ref{Data_Eng}. Following data pre-processing and refinement, models are subsequently implemented through learning and decoding in the Python programming language and facilitated by \textit{pomegranate} \citep{Schreiber2016}, by utilizing a 56-core workstation running at 2.60 GHz with 192 GB of RAM.

\begin{center}
\begin{longtable}{|p{.3\textwidth} X | p{.35\textwidth} X | p{.3\textwidth} X |}
\caption{List of features for \textit{Intra-State} HMMs in UTFM.} \label{tab:IntraStateFeatures}\\

\hline 	\textbf{\textit{Intra-State} HMM for UTFM}  & \textbf{Hidden States} & \textbf{Observations}\\ \hline 
\endfirsthead

\multicolumn{3}{c}%
{{\bfseries \tablename\ \thetable{} -- continued from previous page}} \\
\hline 	\textbf{\textit{Intra-State} HMM for UTFM}  & \textbf{Hidden States} & \textbf{Observations}\\ \hline 
\endhead

\hline \multicolumn{3}{|r|}{{Continued on next page}} \\ \hline
\endfoot

\hline \hline
\endlastfoot
TAS (Turnaround Schedule)  & \textit{SWAP\_FLT\_FLAG, SCHED\_ACFT\_TYPE, SCHED\_TURN\_MINS, tod\_sched\_PB} & RTE, FREQ, PAX DMD\\ \hline
TOS (Taxi-out Schedule)  & \textit{taxi\_out, tod\_actl\_TO, sched\_block\_mins} & RTE, FREQ, PAX DMD\\ \hline
ES (Enroute Schedule)   & \textit{actl\_enroute\_mins, tod\_actl\_LD, sched\_block\_mins}  & RTE, FREQ, PAX DMD\\ \hline
TIS (Taxi-in Schedule)  & \textit{taxi\_in, tod\_sched\_GP, sched\_block\_mins} & RTE, FREQ, PAX DMD\\ \hline
TAD (Turnaround Decision)  & \textit{shiftper\_sched\_PB, ADJST\_TURN\_MINS, DELY\_MIN, SWAP\_FLT\_FLAG} & ORIG, DEST, FREQ, PAX DMD, DISRP\\ \hline
TOD (Taxi-out Decision)  & \textit{late\_out\_vs\_sched\_mins, shiftper\_actl\_PB, DELY\_MIN} & ORIG, DEST, FREQ, PAX DMD, DISRP\\ \hline
ED (Enroute Decision) & \textit{shiftper\_actl\_TO, shiftper\_actl\_LD, DOT\_DELAY\_MINS} & ORIG, DEST, FREQ, PAX DMD, DISRP\\ \hline
TID (Taxi-in Decision)  & \textit{DOT\_DELAY\_MINS, shiftper\_sched\_GP, shiftper\_actl\_GP} & ORIG, DEST, FREQ, PAX DMD, DISRP\\ \hline
TAO (Turnaround Outcome)  & \textit{SWAP\_FLT\_FLAG, ACTL\_ACFT\_TYPE, ACTL\_TURN\_MINS, tod\_actl\_PB} & RTE, FREQ, PAX DMD\\ \hline
TOO (Taxi-out Outcome) & \textit{taxi\_out, tod\_actl\_TO, actl\_block\_mins} & RTE, FREQ, PAX DMD\\ \hline
EO (Enroute Outcome)  & \textit{actl\_enroute\_mins, tod\_actl\_LD, actl\_block\_mins}  & RTE, FREQ, PAX DMD\\ \hline
TIO (Taxi-in Outcome)  & \textit{taxi\_in, tod\_actl\_GP, actl\_block\_mins} & RTE, FREQ, PAX DMD\\
\end{longtable}
\end{center}

\begin{center}
\begin{longtable}{|p{.3\textwidth} X | p{.35\textwidth} X | p{.3\textwidth} X |}
 \caption{List of features for \textit{Inter-State} HMMs in UTFM.}\label{tab:InterStateFeatures}\\

\hline  \textbf{\textit{Inter-State} HMM for UTFM}  & \textbf{Hidden States} & \textbf{Observations}\\ \hline 
\endfirsthead

\multicolumn{3}{c}%
{{\bfseries \tablename\ \thetable{} -- continued from previous page}} \\
\hline 	\textbf{\textit{Inter-State} HMM for UTFM} & \textbf{Hidden States} & \textbf{Observations}\\ \hline 
\endhead

\hline \multicolumn{3}{|r|}{{Continued on next page}} \\ \hline
\endfoot

\hline \hline
\endlastfoot
	 TAS $\rightarrow$ TOS  & \textit{SWAP\_FLT\_FLAG, SCHED\_ACFT\_TYPE, SCHED\_TURN\_MINS, tod\_sched\_PB} & \textit{taxi\_out, tod\_actl\_TO, sched\_block\_mins}\\ \hline
	 TOS $\rightarrow$ ES  & \textit{taxi\_out, tod\_actl\_TO, sched\_block\_mins} & \textit{actl\_enroute\_mins, tod\_actl\_LD, sched\_block\_mins}\\ \hline
	 ES $\rightarrow$ TIS   & \textit{actl\_enroute\_mins, tod\_actl\_LD, sched\_block\_mins}  & \textit{taxi\_in, tod\_sched\_GP, sched\_block\_mins}\\ \hline
	 TAD $\rightarrow$ TOD  & \textit{shiftper\_sched\_PB, ADJST\_TURN\_MINS, DELY\_MIN, SWAP\_FLT\_FLAG} & \textit{late\_out\_vs\_sched\_mins, shiftper\_actl\_PB, DELY\_MIN}\\ \hline
	 TOD $\rightarrow$ ED  & \textit{late\_out\_vs\_sched\_mins, shiftper\_actl\_PB, DELY\_MIN} & {shiftper\_actl\_TO, shiftper\_actl\_LD, DOT\_DELAY\_MINS}\\ \hline
	 ED $\rightarrow$ TID  & \textit{shiftper\_actl\_TO, shiftper\_actl\_LD, DOT\_DELAY\_MINS} &   \textit{DOT\_DELAY\_MINS, shiftper\_sched\_GP, shiftper\_actl\_GP}\\ \hline
	 TAO $\rightarrow$ TOO & \textit{SWAP\_FLT\_FLAG, ACTL\_ACFT\_TYPE, ACTL\_TURN\_MINS, tod\_actl\_PB}  &  \textit{taxi\_out, tod\_actl\_TO, actl\_block\_mins}\\ \hline
	 TOO $\rightarrow$ EO  &  \textit{taxi\_out, tod\_actl\_TO, actl\_block\_mins}& {actl\_enroute\_mins, tod\_actl\_LD, actl\_block\_mins}\\ \hline
	 EO $\rightarrow$ TIO  & \textit{actl\_enroute\_mins, tod\_actl\_LD, actl\_block\_mins}  & \textit{taxi\_in, tod\_actl\_GP, actl\_block\_mins} \\ \hline
	 TAS $\rightarrow$ TAD & \textit{SWAP\_FLT\_FLAG, SCHED\_ACFT\_TYPE, SCHED\_TURN\_MINS, tod\_sched\_PB} &  \textit{shiftper\_sched\_PB, ADJST\_TURN\_MINS, DELY\_MIN, SWAP\_FLT\_FLAG}\\ \hline
	 TOS $\rightarrow$ TOD  & \textit{taxi\_out, tod\_actl\_TO, sched\_block\_mins}  & \textit{late\_out\_vs\_sched\_mins, shiftper\_actl\_PB, DELY\_MIN} \\ \hline
	 ES $\rightarrow$ ED  & \textit{actl\_enroute\_mins, tod\_actl\_LD, sched\_block\_mins} & \textit{shiftper\_actl\_TO, shiftper\_actl\_LD, DOT\_DELAY\_MINS} \\ \hline
	 TIS $\rightarrow$ TID  & \textit{taxi\_in, tod\_sched\_GP, sched\_block\_mins}  & \textit{DOT\_DELAY\_MINS, shiftper\_sched\_GP, shiftper\_actl\_GP}\\ \hline
	 TAD $\rightarrow$ TAO  & \textit{shiftper\_sched\_PB, ADJST\_TURN\_MINS, DELY\_MIN, SWAP\_FLT\_FLAG} & \textit{SWAP\_FLT\_FLAG, ACTL\_ACFT\_TYPE, ACTL\_TURN\_MINS, tod\_actl\_PB}\\ \hline
	 TOD $\rightarrow$ TOO  & \textit{late\_out\_vs\_sched\_mins, shiftper\_actl\_PB, DELY\_MIN}  & \textit{taxi\_out, tod\_actl\_TO, actl\_block\_mins}\\ \hline
	 ED $\rightarrow$ EO  & \textit{shiftper\_actl\_TO, shiftper\_actl\_LD, DOT\_DELAY\_MINS} & \textit{actl\_enroute\_mins, tod\_actl\_LD, actl\_block\_mins}\\ \hline
	 TID $\rightarrow$ TIO  & \textit{DOT\_DELAY\_MINS, shiftper\_sched\_GP, shiftper\_actl\_GP} & \textit{taxi\_in, tod\_actl\_GP, actl\_block\_mins} \\
\end{longtable}
\end{center}

\subsection{UTFM Input and Output Features}
Table \ref{tab:IntraStateFeatures} and Table \ref{tab:InterStateFeatures} reveal the hidden states (latent output data features) and observations (observed input data features) for all \textit{Intra-State} and \textit{Inter-State} HMMs, respectively, that constitute an aggregate HMM which defines the UTFM. The selection of specific hidden and observation data features, for all \textit{Intra-State} and \textit{Inter-State} HMMs that define the UTFM, was informed partly by literature \citep{Clarke1998, Midkiff2004, Hao2013}, exploratory data analysis \citep{Ogunsina2021}, and partly by discussions with human experts at the AOCC of the U.S. airline that provided the raw historical data. We adopted this hybrid feature selection approach to ensure that data features which are appropriately relevant at a specific phase of activity in the UTFM are parameters of the corresponding HMM that represents that phase of activity for airline disruption management. 

For \textit{Intra-State} HMMs listed in Table \ref{tab:IntraStateFeatures}, observations (i.e. observed aleatoric data features) are defined by data features that are strictly subject to aleatoric uncertainty with respect to how often they are considered, by human specialists, in order to attain optimal schedules during the airline scheduling process shown in Fig.~\ref{fig:AirlineScheduling}. Therefore, observations for \textit{Intra-State} HMMs, listed in Table \ref{tab:IntraStateFeatures}, include data features that represent the following: origin airport location and flight origin (ORIG), destination airport location (DEST), flight operating period in a calendar year (FREQ), route distance between origin and destination airports (RTE), number of passengers available for flight (PAX DMD), and random disruption types such as inclement weather (DISRP). ORIG, DEST, FREQ, RTE, and PAX DMD represent determinate aleatoric features that are determined by the airline, which are subject to aleatoric uncertainty at all phases of activity in the UTFM. As such, these features are indicative of the uniqueness of a particular flight schedule with respect to the airline route network. DISRP represents indeterminate aleatoric features that are subject to uncertainty which can not be readily controlled by an airline, and thus represent pure aleatory in airline disruption management. Hidden states (i.e. epistemic output data features) for \textit{Intra-State} HMMs represent data features that are strictly subject to epistemic uncertainty with respect to the concatenation (interaction) of latent data features with the highest probability of occurrence, which indicate the activity patterns of human specialists (i.e. decision-making) for attaining optimal schedules during the airline scheduling process.

\begin{figure}[ht!]
	\centering
	\includegraphics[width=0.95\textwidth]{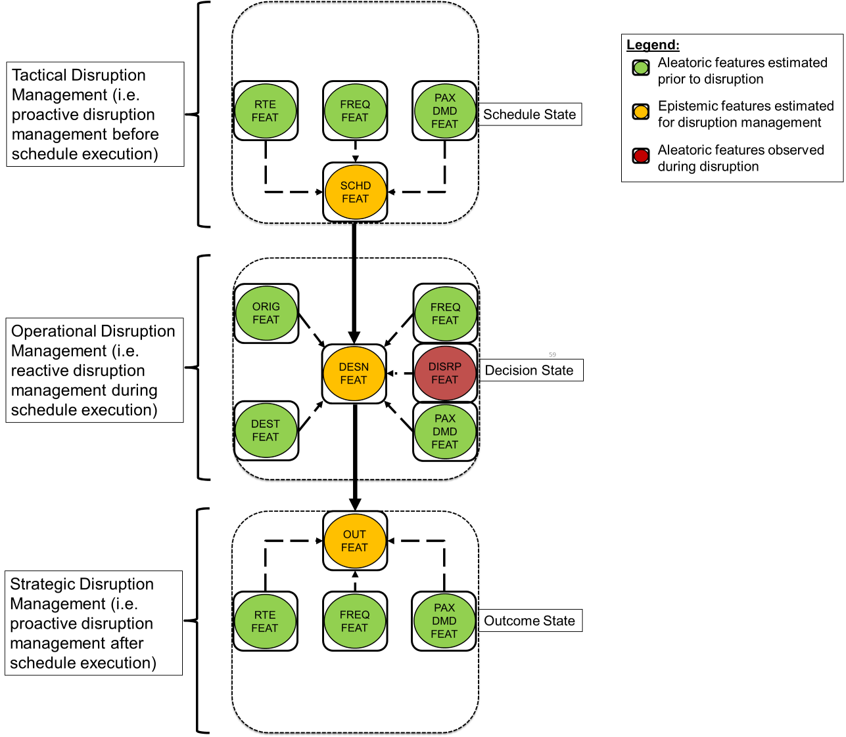}
	\caption{Phases of disruption management with respect to schedule execution}
	\label{fig:LearningPhases}
\end{figure}

For \textit{Inter-State} HMMs listed in Table \ref{tab:InterStateFeatures}, observations (i.e. observed epistemic data features) represent data features that are strictly subject to epistemic uncertainty with respect to the Viterbi probability (i.e. probability of the most likely sequence of latent data features estimated by an \textit{Intra-State} HMM) at an immediate future phase of activity in the UTFM, while hidden states (i.e. latent epistemic data features) represent data features whose concatenations are strictly subject to epistemic uncertainty with respect to the Viterbi probability estimated by a characteristic \textit{Intra-State} HMM in the present phase of activity in the UTFM during airline schedule planning and disruption management.

\subsection{UTFM Learning}
\subsubsection{Defining Hidden States and Observations}
Fig.~\ref{fig:LearningPhases} reveals a one-dimensional spatiotemporal representation of the UTFM reduced along the operation sequence axis (i.e. arbitrary column in Fig.~\ref{fig:UncertaintyModel}). Yellow plates, indicated by SCHD FEAT, DESN FEAT, and OUT FEAT in Fig.~\ref{fig:LearningPhases}, are representative of epistemic data features which define separate hidden states for \textit{Intra-State} HMMs at each phase of flight operation along the operation sequence axis (i.e. Turnaround, Taxi-Out, Enroute, and Taxi-In) in the UTFM detailed in Fig.~\ref{fig:PGM_UTFM}. In that regard, SCHD FEAT represents data features that define hidden states for TAS, TOS, ES, and TIS \textit{Intra-State} HMMs in the UTFM; DESN FEAT represents data features that define hidden states for TAD, TOD, ED, and TID \textit{Intra-State} HMMs, while OUT FEAT is representative of data features that define hidden states for TAO, TOO, EO, and TIO states in the UTFM. Green and red plates in Fig.~\ref{fig:LearningPhases} are representative of uncertainty from determinate and indeterminate and aleatoric features for disruption management respectively, which define observations (inputs) for all \textit{Intra-State} HMMs in the UTFM. 

\subsubsection{Data Segmentation for Learning}
We employ the two separate lots of data in the full data set, defined as the \textit{non-disrupted} and \textit{disrupted} data sets, to learn optimal parameters of all HMMs that define different phases of activity for disruption management in the UTFM. The \textit{non-disrupted} data set contains six hundred and twenty thousand instances of flight schedules in the airline network that executed as originally planned between September 2016 and September 2017. As such, the \textit{non-disrupted} data set contains appropriate latent (hidden) and observation data features for flight schedules that executed without any uncertainty from indeterminate aleatoric features (i.e. random disruption features). Thus, we use the \textit{non-disrupted} data set to calibrate \textit{Intra-State} HMMs that define the tactical and strategic (i.e. Schedule and Outcome) phases of activity for disruption management in the UTFM. Unlike the \textit{non-disrupted} data set, the \textit{disrupted} data set contains all instances of flight schedules that executed through irregular operations due to delays in the airline route network from September 2016 to September 2017. Hence, the \textit{disrupted} data set comprises of instances of flight schedules that executed with uncertainty from indeterminate aleatoric features over a one year period for separate functional roles in the AOCC. 

Therefore, we conduct \textit{Intra-State} HMM learning for operational disruption management (i.e. Decision activity phases in the UTFM) by utilizing the \textit{disrupted} data set. Similarly, we also utilize the \textit{disrupted} data set to learn the optimal parameters of all \textit{Inter-State} HMMs along the operation sequence and schedule-change sequence axes in the UTFM for separate functional roles in the AOCC. To demonstrate the application of the UTFM in this paper, we only consider disruptions due to weather-related events, because irregular operations due to weather disruptions affect all problem dimensions during airline disruption management. As such, the \textit{non-disrupted} data set is used to calibrate the \textit{Intra-State} HMMs for tactical and strategic disruption management; a \textit{disrupted} data set, with over twelve thousand instances of delayed flight schedules due to weather-related disruptions, is used to calibrate the \textit{Intra-State} HMMs for operational disruption management and all \textit{Inter-State} HMMs respectively. Instances of delayed flight schedules used for training represent 99\% of the complete \textit{disrupted} data set, while the remaining 1\% is used later in this paper as new (disrupted) flight schedules or unseen test data to illustrate the UTFM. All \textit{disrupted} and \textit{non-disrupted} data sets used for training and validation are instantiated and segmented by using a random seed of 42 to ensure reproducible models.

\subsubsection{Learning and Validation}
\textit{Intra-State} and \textit{Inter-State} HMM learning for the development of the UTFM is implemented first by fitting data feature samples for hidden states to standard normal probability distributions that define the components of the initial measure of the UTFM. Next, samples (set) of observed data features are grouped as observations and the initial HMM state transition parameters are set as uniform distributions based upon the total number of hidden states, before invoking the Baum-Welch algorithm set to a convergence criterion of $1e^{-9}$. This ensures that UTFM learning executes in polynomial time. We perform a 5-fold cross-validation \citep{Kohavi1995} of Baum-Welch training on the sets of observations by examining marginal probability distributions of latent states across different folds to ensure modeling uniformity and generalizability, for approbation of a candidate optimal \textit{Intra-State} or \textit{Inter-State} HMM trained on the complete set of observations. The cross-validation technique is used to assess the performance (goodness) of a trained HMM (for the UTFM) for estimating the likelihood of new observation (input) data, by verifying that the sums of the log likelihood of an appropriate test set of observations across each of the five folds and corresponding state probability distributions are consistent \citep{Ogunsina2019b}. Adopting the 5-fold model cross-validation for assessing the goodness of all HMMs during \textit{Intra-State} and \textit{Inter-State} learning revealed that the percentage error from the mean of the total sum of the log-likelihoods of suitable test observation data across all five folds was less than 1\%. This indicates that the model performance observed for each fold is consistent across all folds during \textit{Intra-State} and \textit{Inter-State} HMM learning.

\subsection{UTFM Decoding}
\begin{figure}[b!]
	\centering
	\includegraphics[width=0.99\textwidth]{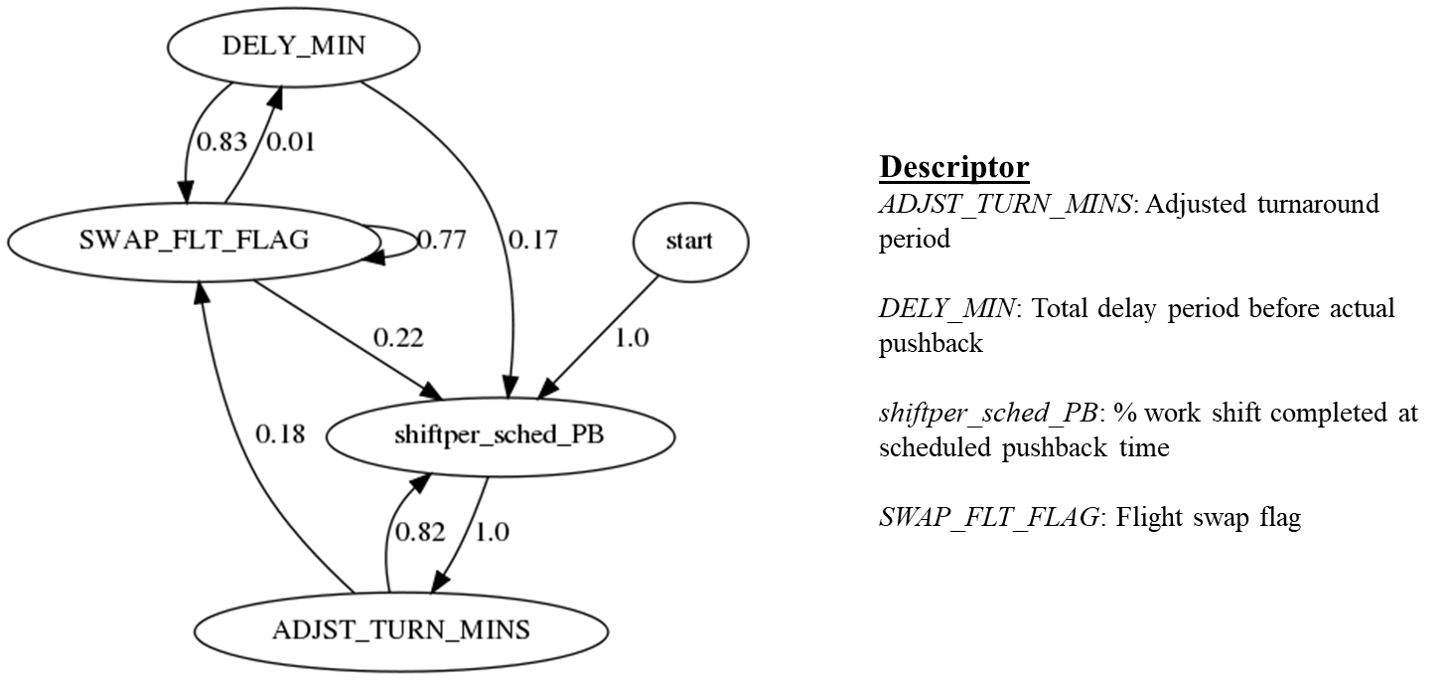}
	\caption{State transition graph of optimal \textit{Intra-State} HMM for remaining in turnaround decision (TAD)}
	\label{fig:TAD_IntraState}
\end{figure}
Upon utilizing refined training data to learn the optimal parameters for \textit{Intra-State} and \textit{Inter-State} HMMs, a hidden Markov model (i.e. probabilistic finite state machine) representation of the UTFM is assembled to enable the decoding of new (unseen) data that represent disrupted flight schedules, by setting the weighted estimates of Viterbi probabilities estimated from all \textit{Intra-State} and \textit{Inter-State} HMMs as parameters of the aggregate \textit{left-right} HMM that represents the UTFM, before applying the Viterbi algorithm to decode (predict) the most likely sequence of state components (i.e. phases of activity during airline disruption management) in the UTFM due to observed inputs from a specific disrupted flight schedule.

\subsubsection{\textit{Intra-State} HMM Decoding}
Fig.~\ref{fig:TAD_IntraState} shows an optimal state transition graph for hidden state features from a trained \textit{Intra-State} HMM for remaining in the Turnaround Decision (TAD) phase of activity in the UTFM. Based upon the graph shown in Fig.~\ref{fig:TAD_IntraState}, a specialist agent will commence decision-making for the turnaround phase of activity in the UTFM for operational disruption management first by assessing how much time there is until the scheduled aircraft pushback time, before considering (transitioning) to adjust the aircraft turnaround time (i.e. start probability of 1 and transition probability of 1). In the less likely event that the specialist agent does not return to assessing the time remaining prior to the scheduled aircraft pushback, a consideration to swap the aircraft is most likely (transition probability of 0.18) and there is a 77\% likelihood that the process to swap the aircraft type will continue throughout the turnaround phase of flight operation during operational disruption management. Fig.~\ref{fig:TAD_IntraState} reveals that there is almost no prerogative for the specialist agent to consider delaying aircraft pushback time after swapping aircraft during the turnaround phase of flight operation for operational disruption management, as evidenced by the negligible transition probability of 1\%.

\begin{figure}[t!]
	\centering
	\includegraphics[width=0.99\textwidth]{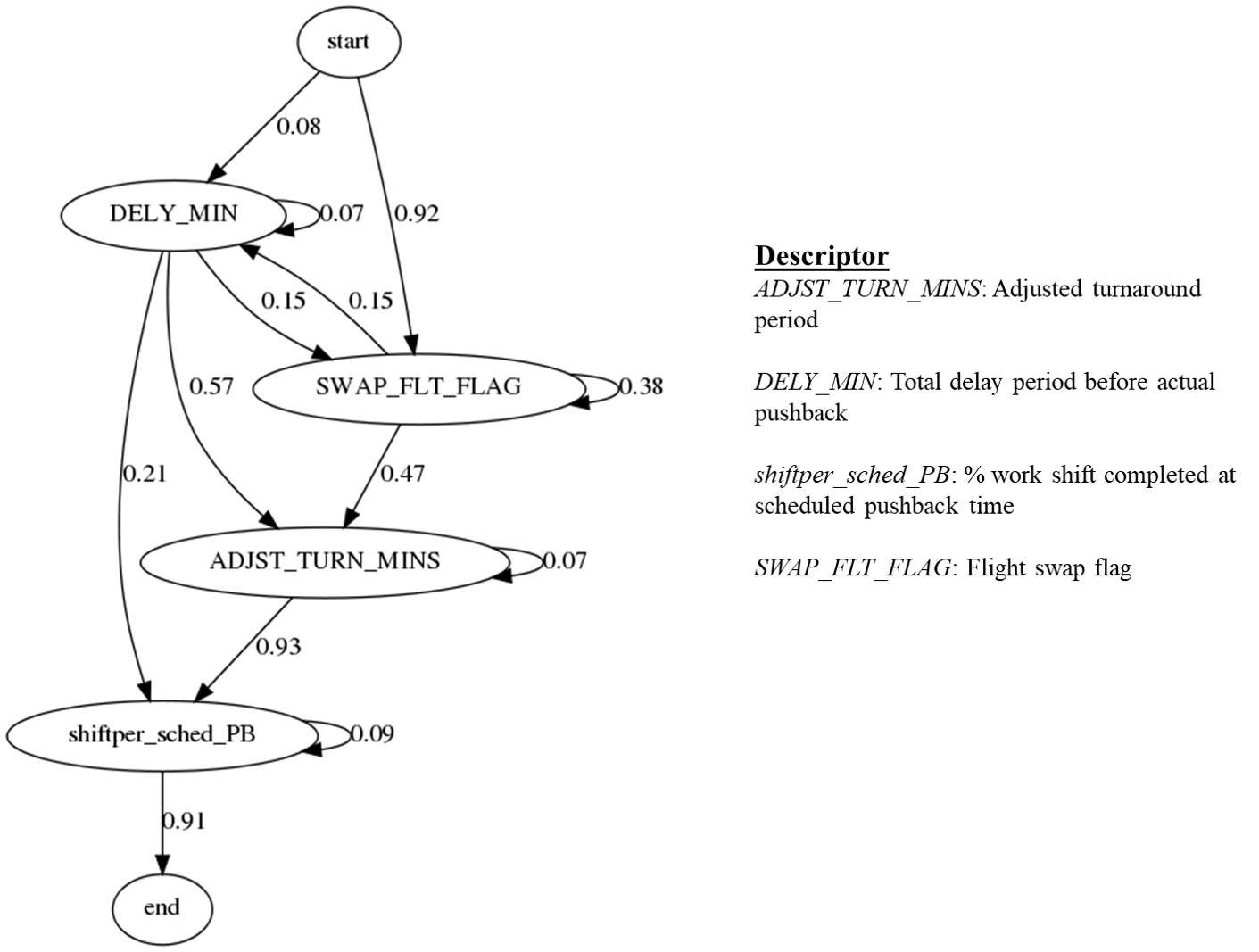}
	\caption{State transition graph of optimal \textit{Inter-State} HMM for transition from turnaround decision (TAD) to turnaround outcome (TAO)}
	\label{fig:TAD_TAO_InterState}
\end{figure}

\subsubsection{\textit{Inter-State} HMM Decoding}
Fig.~\ref{fig:TAD_TAO_InterState} shows an optimal state transition graph for hidden state features from a trained \textit{Inter-State} HMM for transitioning from the Turnaround Decision (TAD) phase of activity to the Turnaround Outcome (TAO) phase of activity in the UTFM. From the graph shown in Fig.~\ref{fig:TAD_TAO_InterState}, a specialist agent will most likely commence the transitioning from operational decision-making for the turnaround phase of activity to strategic (proactive) decision-making for a future turnaround phase of activity in the UTFM for disruption management, first by assessing flight swap (start probability of 0.92 and internal state probability of 0.38), before a most likely transition to consider adjusting aircraft turnaround time (transition probability of 0.47 and internal state probability of 0.07). In the much less likely event that the specialist agent commences the transition to strategic disruption management by considering delay time before pushback first (start probability of 0.08 and internal state probability of 0.07), there is a 57\% likelihood that the decision to adjust the turnaround time will follow, and transitioning for strategic disruption management of the turnaround phase of future flight operation concludes by assessing the work shift (time available) for the next aircraft pushback schedule (end probability of 0.91 and internal state probability of 0.09).

Unlike the ergodic structure of the optimal state transition graph for the TAD \textit{Intra-State} HMM represented in Fig.~\ref{fig:TAD_IntraState}, the optimal state transition graph for the \textit{Inter-State} HMM for transitioning between TAD and TAO phases of activity in the UTFM (depicted in Fig.~\ref{fig:TAD_TAO_InterState}) is modeled as a non-ergodic structure by introducing an absorption state (i.e. `end' state) to characterize a definite transition process between both phases of activity.
Thus, we apply ergodic (and non-ergodic) properties to determine the optimal parameters of all \textit{Intra-State} and \textit{Inter-State} HMMs that constitute different phases of activity in the UTFM. 

\section{UTFM Results} \label{utfm_results}

\begin{figure}[ht!]
	\centering
	\includegraphics[width=0.6\textwidth]{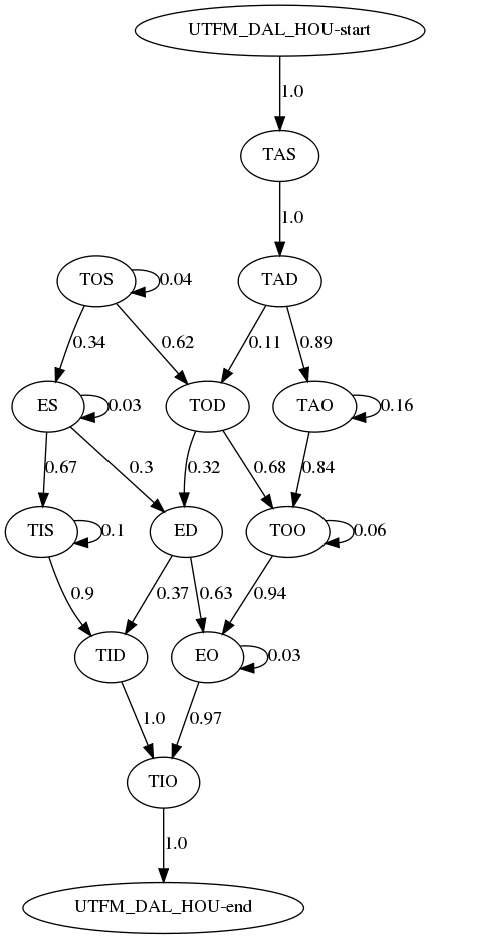}
	\caption{Probabilistic graphical map for UTFM assessment of a specific disrupted DAL-HOU flight}
	\label{fig:UTFM_DAL_HOU}
\end{figure}

We now evaluate two distinct flight schedules, impacted by two different kinds of weather-related disruptions (i.e. uncertainty from indeterminate aleatoric features), which represent two separate samples from \textit{disrupted} test (unseen) data set, by employing the UTFM for airline disruption management. We selected these flight schedules as candidate test subjects for our demonstration because they represent major routes in the network of the U.S. airline carrier that provided the data which enabled the development of the UTFM. For our assessments, we implement an aggregate non-ergodic HMM representation of the UTFM, such that the disruption management process strictly starts at Turnaround Schedule (TAS) phase of activity and ends at Taxi-In Outcome (TIO) phase of activity. 

Fig.~\ref{fig:UTFM_DAL_HOU} shows the probabilistic graphical model representation of the UTFM for disruption management on the operation of a specific flight from Dallas to Houston (DAL-HOU), which was disrupted by air traffic control (ATC) hold for bad weather at Dallas (i.e. \textit{HDO6} delay code). Fig.~\ref{fig:UTFM_DAL_HOU} reveals that there is a 100\% likelihood that a specialist agent transitions to employ reactive disruption management measures from tactical disruption management measures during the turnaround phase of flight operation at Dallas (100\% transition probability from TAS to TAD). As such, to effectively resolve the same disruption instance in the future, the most likely approach is adjust or update features in the turnaround, taxi-out and enroute phases of flight operation accordingly, as evidenced by internal state probabilities of 16\%, 6\%, and 3\% for remaining in the TAO, TOO, and EO phases of activity respectively. Furthermore, Fig.~\ref{fig:UTFM_DAL_HOU} reveals that the tactical disruption management initiative implemented for the turnaround flight phase to address the ATC hold for inclement weather at Dallas for that particular Dallas to Houston flight was ineffective, as evidenced by the lack of transition from the turnaround phase of flight operation to the taxi-out phase of operation (i.e. zero probability of transition from TAS to TOS). As such delays were most likely incurred during the turnaround phase of operation while executing that particular flight from Dallas to Houston. However, tactical initiatives proved somewhat effective during the taxi-out, enroute, and taxi-in phases of activity for disruption management of the Dallas to Houston flight, affirmed by internal state probabilities (i.e. interaction of hidden data features in \textit{Intra-State} HMMs) of 4\%, 3\%, and 10\% for remaining in the TOS, ES, and TIS phases of activity respectively. 

\begin{figure}[t!]
	\centering
	\includegraphics[width=0.6\textwidth]{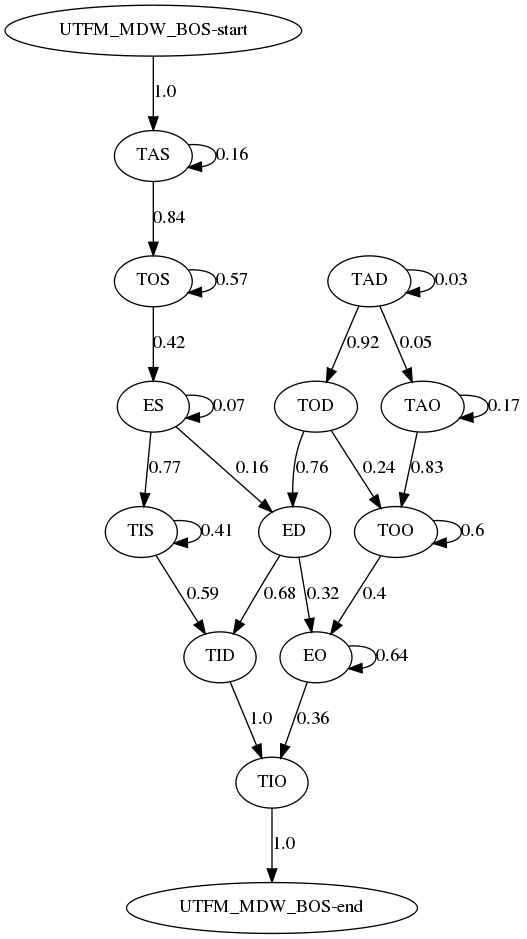}
	\caption{Probabilistic graphical map for UTFM assessment of a specific disrupted MDW-BOS flight}
	\label{fig:UTFM_MDW_BOS}
\end{figure}

Fig.~\ref{fig:UTFM_MDW_BOS} shows the probabilistic graphical model representation of the UTFM for disruption management on the operation of a specific flight from Chicago to Boston (MDW-BOS), which was disrupted by ATC hold for bad weather en route to or at Boston (i.e. \textit{HDO7} delay code). Fig.~\ref{fig:UTFM_MDW_BOS} affirms that it is more likely that the tactical disruption management measures a specialist agent employs for disruption management of bad weather at Boston are proactively effective for the turnaround and taxi-out phases of flight operation, as indicated by internal state probabilities of 0.16 and 0.57 for TAS and TOS respectively and zero likelihood of transitions from those states to TAD and TOD respectively. Even though the tactical disruption management measures for addressing the inclement weather disruption at Boston in the enroute and taxi-in phases of activity are somewhat effective, there may be situations where decision-making for reactive disruption management at the enroute and taxi-in phases of activity during schedule execution may prove useful; as evidenced by the state transition probabilities of 0.16 and 0.59 from ES to ED and TIS to TID respectively. Furthermore, Fig.\ref{fig:UTFM_MDW_BOS} reveals that the proactive tactical disruption management measures for the turnaround and taxi-out phases of operation, implemented prior to departure from Chicago, were optimally effective for resolving ATC delay at Boston, as there are no transitions from TAS to TAD and TOS to TOD phases on activity in the UTFM. As such delays during the flight were accrued at the enroute and taxi-in phases of operation during disruption management. However, the UTFM representation from Fig.\ref{fig:UTFM_MDW_BOS} reveals that strategic disruption management initiatives to improve the future disruption resolution for this particular flight from Chicago to Boston, due to uncontrollable aleatoric uncertainty from inclement weather at Boston, do exist for turnaround, taxi-out and enroute phases of flight operation; as indicated by internal state probabilities of 17\%, 60\%, and 64\% for remaining in the TAO, TOO, and EO phases of activity respectively.

\section{Conclusion} \label{conclusion}
Existing practices for airline disruption management are defined by human-centric methods that do not definitively examine uncertainty in long-term (proactive) and short-term (reactive) scheduling initiatives for mitigating irregular operations during schedule execution. To this end, we introduced and demonstrated a data-driven and modular activity recognition framework that utilizes a unique class of probabilistic graphical models (i.e. the hidden Markov model) to learn and assess pertinent patterns and behaviors for proactive (tactical) disruption management prior to schedule execution, reactive (operational) disruption management during schedule execution and proactive (strategic) disruption management after schedule execution; all of which are necessary for achieving robust airline disruption management. An effective application of two different classes of dynamic programming algorithms, i.e. the Baum-Welch and Viterbi algorithms, were used to respectively learn and decode the parameters of different HMMs that constitute an overarching HMM required for enabling the assessment of two real-world flight schedules from a major U.S. airline network, disrupted due to different weather-related delays during schedule execution.

The implications of the results from the two particular weather-disrupted flight schedules assessed in this paper reveal that disruption resolution measures enforced during phases of flight where the aircraft is on the ground (e.g. turnaround and taxi-in) are tantamount to attaining robust airline disruption management. Decision-making initiatives employed at phases of flight where the aircraft is on the ground are very likely to propagate to the airborne phases of flight operation, consequently shaping the disruption management outlook for a particular disrupted flight. Furthermore, our relational dynamic Bayesian network (RDBN) architecture\textemdash for the assessment of uncertainty transfer between different phases of flight operation and schedule evolution\textemdash proved useful in rationalizing complex interactions of separate drivers for proactive and reactive disruption management at different phases of activity during the airline scheduling process. For air traffic control hold arising from inclement weather at the departure airport, the RDBN (illustrated by Fig.~\ref{fig:UTFM_DAL_HOU}) revealed a severed transition between the turnaround and taxi-out phases of flight during tactical disruption management. Thus, prior to schedule execution, the likelihood of effectively completing a scheduled flight\textemdash given weather-related disruptions at the departure airport during schedule execution\textemdash is sensitive to foresighted disruption management initiatives enacted for turnaround and taxi-out phases of flight. For air traffic control hold originating from inclement weather at the arrival airport, the RDBN (illustrated by Fig.~\ref{fig:UTFM_MDW_BOS}) revealed a complete transition process between all respective phases of flight during tactical disruption management. Hence, given weather-related disruptions at the arrival airport during schedule execution, the likelihood of practically completing a scheduled flight is unlikely to be affected prior to schedule execution.

\section{Limitations and Future Research Direction} \label{future_work}
Although the work presented in this paper introduces a novel data-driven concept and its application for uncertainty quantification and propagation in the airline scheduling process for robust disruption management, there exist a few areas for further research. First, the data used to inform the development of the uncertainty transfer function model (UTFM), based upon our RDBN architecture, was provided by an airline that primarily operates a point-to-point route network structure. As such, there is a need for investigation of an equivalent framework developed based upon data from a major airline that utilizes a hub and spoke route network. Moreover, to facilitate system-wide disruption management measures like the FAA collaborative decision making initiative, readily accessible data from other air transportation system stakeholders (such as airports) can be inculcated to improve the efficacy of the RDBN architecture (UTFM) for disruption management. 

Second, the selection of specific data features for different phases of activity in the construction of the UTFM introduced in this paper is primarily informed by literature and expert inputs of human specialists from one airline, and may contain biases with respect to separate perspectives for different objectives of air transportation stakeholders for system-wide disruption management. As such, proven non-parametric and unsupervised machine learning techniques can be employed to mitigate and validate biases for ensuring a fairly objective selection of features to represent different air transportation system stakeholders for robust disruption management in the national airspace system. Furthermore, the Baum-Welch algorithm presents an inherently sub-optimal unsupervised learning routine for obtaining component HMMs of the UTFM. To that effect, more research to ensure and enhance solution fidelity of unsupervised machine learning methods is most opportune. 

\section*{Acknowledgement}
The authors would like to thank Blair Reeves, Chien Yu Chen, Kevin Wiecek, Jeff Agold, Dave Harrington, Rick Dalton, and Phil Beck, at Southwest Airlines Network Operations Control (SWA-NOC), for their expert inputs in abstracting the data used for this work.

\section*{Conflict of Interest}
All authors have no conflict of interest to report. 

%fac\newpage
\bibliography{mendeley}

\appendix
\newpage
\section{Nomenclature for determinate aleatoric data features}\label{table:detalea_table}
\begin{longtable}{|p{.3\textwidth} X | p{.40\textwidth} X | p{.30\textwidth} X |}
  \hline
	\textbf{Aleatoric Data Feature} & \textbf{Description} & \textbf{Observation Input Category}\\ \hline
	\textit{dow} & Day of the week & FREQ\\ \hline
	\textit{doy} & Day of the year & FREQ\\ \hline
	\textit{dest\_x\_dir} & Destination airport location in spherical X coordinate & DEST\\ \hline
	\textit{dest\_y\_dir} & Destination airport location in spherical Y coordinate & DEST\\ \hline
	\textit{dest\_z\_dir} & Destination airport location in spherical Z coordinate & DEST\\ \hline
	\textit{moy} & Month of the year & FREQ\\ \hline
	\textit{ONBD\_CT} & Total number of passengers onboard flight  & PAX DMD\\ \hline
	\textit{orig\_x\_dir} & Origin airport location in spherical X coordinate & ORIG\\ \hline
	\textit{orig\_y\_dir} & Origin airport location in spherical Y coordinate & ORIG\\ \hline
	\textit{orig\_z\_dir} & Origin airport location in spherical Z coordinate & ORIG\\ \hline
	\textit{route} & Spherical distance between origin and destination airports & RTE\\ \hline
	\textit{sched\_route\_originator\_flag} &  Flag to indicate first flight of the day & ORIG\\ \hline
	\textit{season} &  Season of the year & FREQ\\ \hline
\end{longtable}

\newpage
\section{Nomenclature for indeterminate aleatoric features}\label{table:indetalea_table}
\begin{longtable}{|p{.25\textwidth} X | p{.25\textwidth} X | p{.25\textwidth} X |  p{.25\textwidth} X |}
  \hline
	\textbf{Aleatoric Data Feature} & \textbf{Description} & \textbf{Observation Input Category} & \textbf{Functional Role} \\ \hline
	\textit{HD03} & Weather holding  & DISRP & Weather\\ \hline
	\textit{HD06} & ATC gate hold for weather at departure station  & DISRP & Weather\\ \hline
	\textit{HD07} &  ATC gate hold for weather at enroute or at destination station  & DISRP & Weather\\ \hline
	\textit{HD08} & Ice on wings / cold-soaked fuel  & DISRP & Weather\\ \hline
	\textit{HD09} & Deicing at gate & DISRP  & Weather\\ \hline
	\textit{MX05} & Inspection due to lightning strike  & DISRP & Weather\\ \hline
	\textit{MX07} & Inspection due to turbulence & DISRP & Weather\\ \hline
	\textit{MXO8} & Hail ice, or snow damage  & DISRP & Weather\\ \hline
\end{longtable}

\newpage
\section{Nomenclature for Epistemic Data Features}\label{table:epis_table}
\begin{longtable}{|p{.3\textwidth} X | p{.40\textwidth} X | p{.30\textwidth} X |}
  \hline
	\textbf{Epistemic Data Feature} & \textbf{Description} & \textbf{Activity Phase in UTFM}\\ \hline
	\textit{ACTL\_ACFT\_TYPE} & Actual aircraft type used & TAO\\ \hline
	\textit{actl\_block\_mins} & Actual blocktime period & TOO, EO, TIO \\ \hline
	\textit{actl\_enroute\_mins} & Actual flight period in the air & EO\\ \hline
	\textit{ACTL\_TURN\_MINS} & Actual turnaround period & TAO\\ \hline
	\textit{ADJST\_TURN\_MINS} & Adjusted turnaround period & TAD\\ \hline
	\textit{DELY\_MINS} & Total delay period before actual pushback & TAD, TOD\\ \hline
	\textit{DOT\_DELAY\_MINS} & Total arrival delay & ED, TID\\ \hline
	\textit{late\_out\_vs\_sched\_mins} & Total departure delay & TOD\\ \hline
	\textit{SCHED\_ACFT\_TYPE} &  Scheduled aircraft type used & TAS\\ \hline
	\textit{sched\_block\_mins} &  Scheduled blocktime period & TOS, ES, TIS\\ \hline
	\textit{SCHED\_TURN\_MINS} &  Scheduled turnaround period & TAS\\ \hline
	\textit{shiftper\_actl\_GP} &  \% work shift completed at actual gate parking time & TID\\ \hline
	\textit{shiftper\_actl\_LD} &  \% work shift completed at actual landing time & ED\\ \hline
	\textit{shiftper\_actl\_PB} &  \% work shift completed at actual pushback time & TOD\\ \hline
	\textit{shiftper\_actl\_TO} &  \% work shift completed at actual takeoff time & ED\\ \hline
	\textit{shiftper\_sched\_GP} &  \% work shift completed at scheduled gate parking time & TID\\ \hline
	\textit{shiftper\_sched\_PB} &  \% work shift completed at scheduled pushback time & TAD\\ \hline
	\textit{SWAP\_FLT\_FLAG} & Flight swap flag & TAS, TAD, TAO\\ \hline
	\textit{taxi\_in} & Taxi-in period & TIS, TIO\\ \hline
	\textit{taxi\_out} & Taxi-out period & TOS, TOO\\ \hline
	\textit{tod\_actl\_GP} &  Actual aircraft gate parking time at destination & TIO\\ \hline
	\textit{tod\_actl\_LD} &  Actual aircraft landing time at destination & EO\\ \hline
	\textit{tod\_actl\_PB} & Actual aircraft pushback time at origin & TAO\\ \hline
	\textit{tod\_actl\_TO} & Actual aircraft takeoff time at origin & TOO\\ \hline
	\textit{tod\_sched\_GP} &  Scheduled aircraft gate parking time at destination & TIS\\ \hline
	\textit{tod\_sched\_PB} &  Scheduled aircraft pushback time at origin & TAS\\ \hline
\end{longtable}

\newpage
\section{Dynamic Programming Algorithm 1}\label{alg:Baum-Welch}
\begin{algorithm}
\caption{\textbf{\textit{Baum-Welch Algorithm}}}
\begin{algorithmic}[1]
\Procedure{BaumWelch}{$Y, X$}
\State $A, B, \alpha, \beta \in Y$ 
\For{$t = 1:N$} 
\State $\gamma(:,t) = \displaystyle\frac{\alpha(:,t)\odot \beta(:,t)}{\sum(\alpha(:,t)\odot \beta(:,t))}$
\newline
\State $\xi(:,:,t) = \displaystyle\frac{(\alpha(:,t)\odot A(t+1))*(\beta(;,t+1)
\odot B(X_{t+1}))^{T}}{\sum(\alpha(:,t)\odot \beta(:,t))}$
\EndFor \Comment{where $N = \lvert X \rvert$ }
\newline
\State $\hat{\pi} = \displaystyle\frac{\gamma(:,1)}{\sum(\gamma(:,1))}$
\newline
\For{$j = 1:K$}
\State $\hat{A}(j,:) = \displaystyle\frac{\sum(\xi(2:N,j,:),1)}{\sum(\sum(\xi(2:N,j,:),1),2)}$
\newline
\State $\hat{B}(j,:) = \displaystyle\frac{X(:,j)^{T}\gamma}{\sum(\gamma,1)}$
\EndFor \Comment{where $K$ is number of states }
\State $\textbf{return} \quad \hat{\pi}, \hat{A}, \hat{B}$
\EndProcedure
\end{algorithmic}
\end{algorithm}

\newpage
\section{Dynamic Programming Algorithm 2}\label{alg:Viterbi}
\begin{algorithm}
\caption{\textbf{\textit{Viterbi Algorithm}}}
\begin{algorithmic}[1]
\Procedure{Viterbi}{$Y, X$}
\State $A, B, \pi \in Y$
\State $\text{Initialize:} \quad \delta_{1} = \pi \circ B_{X_{1}}, \quad a_{1} = 0 $
\For {$t = 2:N$}
\For {$j = 1:K$}
\State $[a_{t}(j), \delta_{t}(j)] = \max_{i}(\log{\delta_{t-1}(:)} + \log{A_{ij}} + \log{B_{X_{i}}(j))} $
\EndFor   \Comment{where $K$ is number of states }
\EndFor  \Comment{where $N = \lvert X \rvert$ }
\State $Z^{*}_{N} = \argmax{\delta_{N}}$
\For {$t=N-1:1$}
\State $Z^{*}_{t}=a_{t+1}Z^{*}_{t+1}$
\EndFor
\State $\textbf{return} \quad Z^{*}_{1:N}$
\EndProcedure
\end{algorithmic}
\end{algorithm}

\vfill
\section{Dynamic Programming Algorithm 3}\label{alg:UTFMlearning}
\begin{algorithm}
\caption{\textbf{\textit{UTFM Learning Algorithm}}}
\begin{algorithmic}[1]
\Procedure{UTFMlearning}{$X, Y$}
    \State{$X_{S} = \{s_{1}, ... , s_{m}\}$, $X_{D} = \{d_{1}, ... , d_{m}\}$, $X_{O} = \{o_{1}, ... , o_{m}\}$}\Comment{Disrupted flight data}
    \ForAll{$j \in (1, 2, ..., m)$} 
        \State{$\mathcal{S^{\prime}} \leftarrow S_{j} $, \quad $\mathcal{D^{\prime}} \leftarrow D_{j} $, \quad $\mathcal{O^{\prime}} \leftarrow O_{j} $, \newline 
        $A^{\prime} \leftarrow \alpha_{ij}: S_{i} \rightarrow S_{j} $, \quad $B^{\prime} \leftarrow \beta_{ij}: D_{i} \rightarrow D_{j} $, \quad $\Gamma^{\prime} \leftarrow \gamma_{ij}: O_{i} \rightarrow O_{j}$, \newline
        $K^{\prime} \leftarrow \kappa_{j}: S_{j} \rightarrow D_{j} $, \quad $\Lambda^{\prime} \leftarrow \lambda_{j}: D_{j} \rightarrow O_{j} $}\Comment{for $i = j-1$ and $i > 0$}
    
        \State{$\mathcal{M^{\prime}} \leftarrow \{\mathcal{S}^{\prime},  \mathcal{D}^{\prime}, \mathcal{O}^{\prime}, A^{\prime}, B^{\prime}, \Gamma^{\prime}, K^{\prime}, \Lambda^{\prime}\}$}\Comment{Initialize Optimal HMM sets for UTFM}
    
        \While{$\lvert X_{S} \rvert, \lvert X_{D} \rvert, \lvert X_{O} \rvert > m$ \quad $\textbf{or}$ \quad $\lnot \mathcal{M^{\prime}}$}
    
            \State{$Y_{S} = \{y_{s1}, ... , y_{sl}\}$, $Y_{D} = \{y_{d1}, ... , y_{dl}\}$, $Y_{O} = \{y_{o1}, ... , y_{ol}\}$}\Comment{Training (flight schedule) data for $l \geq m$}
    
            \State{$S^{\prime}_{j} \leftarrow \Call{BaumWelch}{S_{j},Y_{S}}$, \quad
            $D^{\prime}_{j} \leftarrow \Call{BaumWelch}{D_{j}, Y_{D}}$, \newline
            $O^{\prime}_{j} \leftarrow \Call{BaumWelch}{O_{j}, Y_{O}}$, \quad
            $\alpha^{\prime}_{ij} \leftarrow \Call{BaumWelch}{\alpha_{ij}, Y_{S}}$, \newline
            $\beta^{\prime}_{ij} \leftarrow \Call{BaumWelch}{\beta_{ij}, Y_{D}}$, \quad 
            $\gamma^{\prime}_{ij} \leftarrow \Call{BaumWelch}{\gamma_{ij}, Y_{O}}$, \newline
            $\kappa^{\prime}_{j} \leftarrow \Call{BaumWelch}{\kappa_{j}, Y_{D}}$, \quad
            $\lambda^{\prime}_{j} \leftarrow \Call{BaumWelch}{\lambda_{j}, Y_{O}}$}
    
            \State{$\mathcal{S^{\prime}} \leftarrow S^{\prime}_{j} $, \quad $\mathcal{D^{\prime}} \leftarrow D^{\prime}_{j} $, \quad $\mathcal{O^{\prime}} \leftarrow O^{\prime}_{j} $, \newline 
            $A^{\prime} \leftarrow \alpha^{\prime}_{ij} $, \quad $B^{\prime} \leftarrow \beta^{\prime}_{ij} $, \quad $\Gamma^{\prime} \leftarrow \gamma^{\prime}_{ij}$, \newline $K^{\prime} \leftarrow \kappa^{\prime}_{j}$, \quad $\Lambda^{\prime} \leftarrow \lambda^{\prime}_{j}$}\Comment{Update Optimal HMM sets for UTFM}
        \EndWhile
    \EndFor
    \State{$\mathcal{N_{0}} \leftarrow \{\mathcal{S^{\prime}}, \mathcal{D^{\prime}}, \mathcal{O^{\prime}}\}$, \quad$\mathcal{N_{i \rightarrow j}} \leftarrow \{A^{\prime}, B^{\prime},\Gamma^{\prime}\}$, \quad $\mathcal{N_{i \rightarrow i}} \leftarrow \{K^{\prime}, \Lambda^{\prime}\}$}
    \State{$\mathcal{N_{\rightarrow}} \leftarrow \{\mathcal{N_{i \rightarrow i}}, \mathcal{N_{i \rightarrow j}}\}$ \quad \quad}
    \State{$\mathcal{K} \leftarrow (\mathcal{N_{0}}, \mathcal{N_{\rightarrow}})$}\Comment{Optimal (RDBN) Data Architecture for UTFM} 
\EndProcedure
\end{algorithmic}
\end{algorithm}

\newpage
\section{Dynamic Programming Algorithm 4}\label{alg:UTFMdecoding}
\begin{algorithm}
\caption{\textbf{\textit{UTFM Decoding Algorithm}}}
 \begin{algorithmic}[1]
 \Require   $\mathcal{K}$\Comment{Optimal UTFM Architecture}
 \Procedure{UTFMdecoding}{$X$}
\State{$P(s)\leftarrow\Call{Viterbi}{\mathcal{S^{\prime}}, X_{S}}$, \quad $P(d) \leftarrow \Call{Viterbi}{\mathcal{D^{\prime}}, X_{D}}$, \newline 
$P(o) \leftarrow \Call{Viterbi}{\mathcal{O^{\prime}}, X_{O}}$, \quad
$P(\alpha) \leftarrow \Call{Viterbi}{A^{\prime},X_{S}}$, \newline 
$P(\beta) \leftarrow \Call{Viterbi}{B^{\prime},X_{D}}$, \quad $P(\gamma) \leftarrow \Call{Viterbi}{\Gamma^{\prime},X_{O}}$, \newline 
$P(\kappa) \leftarrow \Call{Viterbi}{K^{\prime}, X_{D}}$, \quad $P(\lambda) \leftarrow \Call{Viterbi}{\Lambda^{\prime}, X_{O}}$}\Comment{Unroll $\mathcal{K}$ with disrupted flight information $X$}
\ForAll{$j \in (1,2,...,m)$}
    \State{$\phi_{j} \leftarrow P(s_{j}) + P(\alpha_{ij}) + P(\kappa_{j})$, \newline 
    $\psi_{j} \leftarrow P(d_{j}) +  P(\beta_{ij}) + P(\lambda_{j})$, \newline 
    $\rho_{j} \leftarrow P(o_{j}) + P(\gamma_{ij})$ \Comment{for $i = j-1$ and $i > 0$}}
    \State{$a \leftarrow \frac{P(s_{j})}{\phi_{j}}, 
    \quad b \leftarrow \frac{P(\alpha_{ij})}{\phi_{j}}, 
    \quad c \leftarrow \frac{P(\kappa_{j})}{\phi_{j}}$,\newline 
    $p \leftarrow \frac{P(d_{j})}{\psi_{j}}, 
    \quad q \leftarrow \frac{P(\beta_{ij})}{\psi_{j}}, 
    \quad r \leftarrow \frac{P(\lambda_{j})}{\psi_{j}},$
    \newline
    $u \leftarrow \frac{P(o_{j})}{\rho_{j}}, 
    \quad v \leftarrow \frac{P(\gamma_{ij})}{\rho_{j}}$} \Comment{Stochastic matrix (state probabilities) for UTFM}
\EndFor
\State $N_{0} \leftarrow \{a, p, u\}$, \quad $N_{i \rightarrow j} \leftarrow \{b, q, v\}$, \quad $N_{i \rightarrow i} \leftarrow \{c, r\}$
\State $N_{\rightarrow} \leftarrow \{N_{i \rightarrow i}, N_{i \rightarrow j}\}$ \quad \quad
\State $K \leftarrow (N_{0}, N_{\rightarrow})$ \quad \quad \Comment{UTFM for disrupted flight}  
\State $\textbf{return} \quad K$ 
\EndProcedure
\end{algorithmic}
\end{algorithm}

\end{document}